\documentclass[pdflatex,sn-mathphys-ay]{sn-jnl}
\usepackage[utf8]{inputenc}
\usepackage{textgreek}

\usepackage{graphicx}%
\usepackage{multirow}%
\usepackage{amsmath,amssymb,amsfonts}%
\usepackage{amsthm}%
\usepackage{mathrsfs}%
\usepackage{mathabx}
\usepackage[title]{appendix}%
\usepackage{xcolor}%
\usepackage{textcomp}%
\usepackage{manyfoot}%
\usepackage{booktabs}%
\usepackage{algorithm}%
\usepackage{algorithmicx}%
\usepackage{algpseudocode}%
\usepackage{listings}%
\usepackage{booktabs}
\usepackage{array}
\usepackage{float}
\usepackage{multirow}
\usepackage{bm}
\usepackage{colortbl}
\usepackage{forest}
\usetikzlibrary{shapes.geometric, arrows.meta, positioning}
\usetikzlibrary{trees}
\usepackage{caption}
\usepackage{tabularx}

\usepackage{subcaption}
\usepackage{makecell}
\usepackage{booktabs}
\usepackage{array}

\newcolumntype{C}{>{\centering\arraybackslash}X}

\begin{document}

\title[Article Title]{A Survey on Hypergame Theory: Modelling Misaligned Perceptions and Nested Beliefs for Multi-Agent Systems}

\author[1]{\fnm{Vince} \sur{Trencsenyi}}\email{vince.trencsenyi@rhul.ac.uk}
\author[1]{\fnm{Agnieszka} \sur{Mensfelt}}\email{agnieszka.mensfelt@rhul.ac.uk}
\author[1]{\fnm{Kostas} \sur{Stathis}}\email{kostas.stathis@rhul.ac.uk}

\affil[1]{\orgdiv{Department of Computer Science}, \orgname{Royal Holloway University of London}, \orgaddress{\street{Egham Hill}, \city{Egham}, \postcode{TW200EX}, \state{Surrey}, \country{United Kingdom}}}

%Please provide an abstract of 150 to 250 words. The abstract should not contain any undefined abbreviations or unspecified references. 
\abstract{

Classical game-theoretic models typically assume rational agents, complete information, and common knowledge of payoffs -- assumptions that are often violated in real-world multi-agent systems (MAS) characterised by uncertainty, misaligned perceptions, and nested beliefs. To overcome these limitations, researchers have proposed extensions that incorporate models of cognitive constraints, subjective beliefs, and heterogeneous reasoning. Among these, \textit{hypergame theory} extends the classical paradigm by explicitly modelling agents’ subjective perceptions of the strategic scenario, known as \textit{perceptual games}, in which agents may hold divergent beliefs about the structure, payoffs, or available actions.
We present a systematic review of agent-compatible applications of hypergame theory, examining how its descriptive capabilities have been adapted to dynamic and interactive MAS contexts. We analyse 49 selected studies from cybersecurity, robotics, social simulation, communications, and general game-theoretic modelling. Building on a formal introduction to hypergame theory and its two major extensions -- hierarchical hypergames and the hypergame normal form (HNF) -- we develop agent-compatibility criteria and an agent-based classification framework to assess integration patterns and practical applicability. Our analysis reveals prevailing tendencies, including the prevalence of hierarchical and graph-based models in deceptive reasoning and the simplification of extensive theoretical frameworks in practical applications. We identify structural gaps, including the limited adoption of HNF-based models, the lack of formal hypergame languages, and unexplored opportunities for modelling human–agent and agent–agent misalignment.
By synthesising trends, challenges, and open research directions, this review provides a new roadmap for applying hypergame theory to enhance the realism and effectiveness of strategic modelling in dynamic multi-agent environments.
}

\maketitle

% \tableofcontents

\section{Introduction}

Strategic decision-making is a key issue in many real-world scenarios involving interactions among multiple agents, and as such is a fundamental agentic capability. Since agents in these settings must choose actions based on the actions of others, game theory naturally emerges as the discipline providing formal tools to model, design, and analyse such systems~\citep{Rasmusen1990-RASGAI}. However, real-world environments are often marked by uncertainty, misaligned perceptions, and nested beliefs, which frequently render traditional models inadequate. To address these limitations, researchers have developed frameworks that incorporate cognitive constraints, recursive reasoning, and heterogeneous agents exhibiting bounded rationality~\citep{Harsanyi1967, simon1972, nagel95, camerer2004cognitivehieararchy, DEKEL2015epistemicgametheory}.

Simpler and less expressive models, such as the \textit{k}-level theory and the cognitive hierarchy model~\citep{camerer2004cognitivehieararchy}, acknowledge differences in players’ reasoning capacities but offer limited insight into the structure and underlying nature of these differences. Epistemic game theory provides a formal language for representing player types, belief hierarchies, and uncertainty; however, its predominantly probabilistic models only partially account for the mechanisms that give rise to these cognitive variations~\citep{DEKEL2015epistemicgametheory}.

In contrast, hypergame theory advances the game-theoretic paradigm by explicitly modelling subjective representations of the strategic scenario -- so-called \textit{perceptual games} -- in which agents may disagree about the components of the game: the participating players, the available actions, or the payoff structure~\citep{Bennett1977}. This capability makes hypergames particularly suitable for multi-agent system (MAS) frameworks that aim to emulate human-like reasoning and cognitive sophistication. Such suitability aligns with recent interest in theory of mind (ToM) --
a psychological model of human reasoning that characterises the ability to attribute mental states to oneself and others~\citep{Premack_Woodruff_1978theoryofmind} -- and its integration into agentic artificial intelligence~\citep{rocha2023tomReview}.

Originally developed as an analytical framework for post-hoc conflict analysis, hypergame theory was not designed for dynamic or interactive agent-based systems. As a result, despite its descriptive power and emerging practical applications, its full potential in MAS environments remains underutilised. This systematic review aims to address that gap by identifying existing agent-compatible applications of hypergame theory, analysing patterns of practical integration, and highlighting both structural challenges and emerging opportunities at the intersection of hypergames and multi-agent systems.

We make several contributions. First, we provide a formal introduction to hypergame theory, including its foundational concepts and its two major revisions: hierarchical hypergames and the hypergame normal form (HNF). Building on this foundation, we define agent-compatibility criteria for hypergame models and apply these criteria in a systematic review of 49 selected papers from the hypergame-theoretic literature. We develop an agent-based classification framework to examine how these works integrate hypergame-theoretic models, categorising them by domain and sub-domain, practical applicability, and degree of integration -- from conceptual inspiration through partial implementation to full deployment. We also construct a structured taxonomy of hypergame formalisms, distinguishing multi-level and HNF models, and map their usage across planning, reasoning, learning, and uncertainty modelling tasks. Furthermore, we synthesise trends and usage patterns, showing that while theoretical integrations of hypergames remain prevalent, practical deployments often rely on simplified or flattened models. Finally, we discuss gaps in infrastructure, such as the lack of agent-oriented hypergame modelling languages, and suggest avenues to address these gaps by drawing inspiration from formal languages such as epistemic logic and GDL-III, which are already used in MAS research.

The remainder of this paper is organised as follows. Section~\ref{sec:hgt} reviews the foundations of hypergame theory, including its game-theoretic roots, the formalisation of games, and the development of hierarchical and HNF-based hypergame models. Section~\ref{sec:sysrev} presents the methodology and results of our systematic review, covering publication trends, a taxonomy of hypergame models, and an analysis of their application in practice -- including application domains, integration methodologies, fidelity, and task alignment. Section~\ref{sec:results} discusses the survey results, with particular attention to usage patterns, domain-model associations, and the need for agent-based hypergame languages. Section~\ref{sec:relwork} situates this work within the broader literature, and Section~\ref{sec:conc} concludes with a summary of findings and directions for future research.

\section{Hypergame Theory}
\label{sec:hgt}
In this section, we first establish the foundations of game-theoretic reasoning by introducing classical models of strategic interaction. We then introduce the theory of hypergames as an extension of game theory that relaxes the assumptions of common knowledge and shared perceptions. Finally, we present two major formalisms of hypergame theory: the Hypergame Normal Form (HNF) and Hierarchical Hypergames, both of which have served as theoretical bases for modelling misaligned perceptions and recursive reasoning in multi-agent systems. For the purpose of introducing the relevant concepts, we use the two-player rock-paper-scissors game as a running example -- a well-known, simple setting in which both players must simultaneously choose one of three available actions: Rock, Paper, or Scissors.

\subsection{Game-theoretic Foundations}

While decision theory provides tools to reason about how an individual agent can derive choices, when another decision-maker is involved, the player may have to explicitly reason about how the opponent is trying to solve the same problem. Game theory is better equipped for such analyses, as it deals with conflicts where decisions and their outcomes are interdependent between agents~\citep{myerson1984introduction}.

\subsubsection{Elements of a Game}

~\citep{Rasmusen1990-RASGAI} introduces the key elements of game-theoretic interactions through the acronym \textit{PAPI}: Players, Actions, Payoff and Information. We define these four elements as follows:

\paragraph{Players}
Players are the decision-making entities involved in the strategic interaction. Game-theoretic models typically assume that players are~\citep{myerson1984introduction}:
\begin{itemize}
    \item \textbf{Rational:} they consistently act to maximise their own utility functions;
    \item \textbf{Strategically intelligent:} they understand the structure of the game and anticipate that others behave similarly.
\end{itemize}

\paragraph{Actions}
Actions represent the immediate choices available to each player in the game they are involved with. In contrast, strategies are agents' plans of actions based on the observed interaction history~\citep{OsborneRubinstein1994}. We distinguish between two classes of strategies~\citep{LuceRaiffa1989games}:
\begin{itemize}
    \item \textbf{Pure:} deterministic mappings from states or histories to actions;
    \item \textbf{Mixed:} probability distributions over pure strategies.
\end{itemize}

\paragraph{Payoffs}
Payoffs quantify each player's preferences over possible outcomes. They are typically represented as real-valued utility functions $U_i: A \to \mathbb{R}$, where $A$ is the joint action space. Numerical payoffs enable the use of optimisation and equilibrium analysis~\citep{vonNeumann1944, Gibbons1992}.

\paragraph{Information}
Information concerns players' knowledge at the time of making a decision. This knowledge encompasses a player's image of the other players, the options available to everyone, and the historical moves. We classify games based on three information criteria:
\begin{itemize}
    \item \textbf{Perfect vs. Imperfect Information}: Whether agents can fully observe others' actions~\citep{OsborneRubinstein1994};
    \item \textbf{Complete vs. Incomplete Information}: Whether agents can fully observe others' payoffs~\citep{Gibbons1992};
    \item \textbf{Symmetric vs. Asymmetric Information}: Whether agents can access the same quality and quantity of information~\citep{Rasmusen1990-RASGAI}.
\end{itemize}

\subsubsection{Formalising Games}

A game-theoretic model of rock-paper-scissors can be formally defined as a tuple $G = (N, A, u)$ where:

\begin{itemize}
    \item $N = \{1,2\}$ is the set of players
    \item $A = A_1 \times A_2$ is the set of joint actions, where for each player $i \in N$:
    \begin{itemize}
        \item $A_i = \{\text{Rock}, \text{Paper}, \text{Scissors}\}$ is the action set of player $i$
        \item A pure strategy for player $i$ is an element $a_i \in A_i$
        \item A mixed strategy for player $i$ is a probability distribution $\sigma_i$ over $A_i$
    \end{itemize}
    \item $u = (u_1, u_2)$ are the utility functions where for each player $i \in N$:
    \begin{align*}
        u_i: A \rightarrow \mathbb{R} \text{ defined as: } \\
        u_i(a_i,a_j) = \begin{cases}
            1  & \text{if } (a_i,a_j) \in W_i \\
            0  & \text{if } a_i = a_j \\
            -1 & \text{if } (a_i,a_j) \in L_i
        \end{cases}
    \end{align*}
    where:
    \begin{itemize}
        \item $W_i = \{(\text{Rock},\text{Scissors}), (\text{Paper},\text{Rock}), (\text{Scissors},\text{Paper})\}$ are winning outcomes
        \item $L_i = \{(\text{Rock},\text{Paper}), (\text{Paper},\text{Scissors}), (\text{Scissors},\text{Rock})\}$ are losing outcomes
    \end{itemize}
\end{itemize}

\paragraph{Expected Utility}
When players employ mixed strategies -- probability distributions over their pure actions -- the outcome of the game becomes probabilistic. The expected utility for a player is then calculated as the average payoff, weighted by the likelihood of each joint action occurring under the players’ chosen strategies~\citep{vonNeumann1944}. Formally, given mixed strategies $\sigma_i$ and $\sigma_j$ for players $i$ and $j$, the expected utility for player $i$ is:
\begin{equation*}
    EU_i(\sigma_i,\sigma_j) = \sum_{a_i \in A_i} \sum_{a_j \in A_j} \sigma_i(a_i)\sigma_j(a_j)u_i(a_i,a_j)
\end{equation*}

Rational players are typically assumed to choose strategies that maximise their expected utility, forming the basis of solution concepts such as the mixed-strategy Nash equilibrium.

\paragraph{Nash Equilibrium}
The Nash Equilibrium (NE) is a fundamental solution concept in game theory, capturing the state when neither player can improve their payoff by a unilateral change of strategy~\citep{Nash1950equi}. Formally, a strategy profile $a^*=(a^*_i,a^*_{-i})$ -- with $-i$ denoting the opponent(s) -- is a NE, for each player $i\in N$:
\begin{equation*}
    u_i(a_i^*, a_{-i}^*) \geq u_i(a_i, a_{-i}^*) \quad \forall a_i \in A_i
\end{equation*}

In case of mixed strategies, the Mixed Strategy Nash Eqilibrium (MSNE) is defined analogously: each player’s mixed strategy must be a best response to the mixed strategies of others. Formally:
\begin{equation*}
    EU_i(\sigma^*_i,\sigma^*_{-i}) \geq EU_i(\sigma_i,\sigma^*_{-i}) \quad \forall \sigma_i \in \Delta(A_i),
\end{equation*}
where $\Delta(A_i)$ denotes the set of mixed strategies over $A_i$.\\

In the case of rock-paper-scissors, no pure strategy profile constitutes a Nash equilibrium. However, a unique mixed-strategy Nash equilibrium exists in which each player chooses Rock, Paper, or Scissors with equal probability~\citep{batzilis19rockpaperscissors}.

\subsection{Games of Misperception}

Real-world agents may differ in their abilities to observe and process information, potentially leading to distinct internal models of others and misaligned understandings of the conflict, the available actions, or the associated outcomes. Hypergames provide an explicit model of such perceptions by capturing each agent's subjective interpretation of the situation~\citep{Bennett1980}. Furthermore, hypergame analyses rely on preference relations rather than numerical payoffs, resulting in a qualitative model of desirability. This approach aims to mitigate the challenges of assigning precise numerical values to real-world outcomes.

Hypergames are considered models of bounded rationality~\citep{simon1972,vane2006advances} because they support reasoning in which an agent may perceive an opponent's choice as irrational, even though that choice is rational from the opponent's own perspective~\citep{kovach2015hypergamereview}. Formally, a hypergame is composed of each player's \textit{perceptual game} -- a version of the conflict that represents the player's interpretation -- and is defined as the tuple $H = (N, \{G_i\}_{i \in N})$ where:
\begin{itemize}
    \item $N = \{1,2,\ldots,n\}$ is the set of agents;
    \item For each agent $i \in N$, $G_i$ represents agent $i$'s perceived game, where $G_i = (N_i, A_i, R_i)$ with:
    \begin{itemize}
        \item $N_i \subseteq N$ is the set of agents as perceived by $i$;
        \item $A_i = \bigtimes_{j \in N_i} A_{ij}$ is the joint action space as perceived by $i$, where:
        \begin{itemize}
            \item $A_{ij}$ is $i$'s perception of $j$'s available actions;
        \end{itemize}
        \item $R_i = \{R_{ij}\}_{j \in N_i}$ is the set of preference relations as perceived by $i$, where:
        \begin{itemize}
            \item $R_{ij} \subseteq A_i \times A_j$ is $i$'s perception of $j$'s preference relation;
            \item For outcomes $x,y \in A_i$, $x>y$ indicates that $i$ prefers outcome $x$ to outcome $y$.
        \end{itemize}
    \end{itemize}
\end{itemize}

Note: elements with indices $ii$ denote player $i's$ own interpretation of the game's components.\\

To illustrate misaligned perceptions, consider an asymmetric version of the rock-paper-scissors game modelled as a hypergame in which one agent's action space is constrained. Let $H = (N, \{G_1, G_2\})$, where:

\begin{itemize}
    \item $G_2$ is Agent 1's perceived game:
    \begin{itemize}
        \item $N_1 = \{1,2\}$;
        \item $A_{11} = A_{12} = \{\text{Rock}, \text{Paper}, \text{Scissors}\}$;
        \item $R_{11}=R_{12}=$
        \begin{equation*}
            \{(P,R), (S,P), (R,S)\} > \{(R,R), (P,P), (S,S)\} > \{(S,R), (R,P), (P,S)\};
        \end{equation*}
    \end{itemize}
    \item $G_2$ is Agent 2's perceived game:
    \begin{itemize}
        \item $N_2 = \{1,2\}$;
        \item $A_{21}=A_{22}=\{\text{Rock}, \text{Paper}\}$;
        \item $R_{21}=R_{22}=$
        \begin{equation*}
            \{(R,R), (P,P)\} > \{(R,P), (P,R)\};
        \end{equation*}
    \end{itemize}
\end{itemize}

This example captures a situation in which Scissors is not recognised as an available action by Agent 2, whereas Agent 1 perceives the complete action space. This form of perceptual asymmetry mirrors the classic ``Fall of France'' hypergame analysis~\citep{BennettDando1979}. The resulting discrepancy in perceived game structure can lead to unexpected outcomes -- referred to in hypergame theory as \textit{strategic surprise}~\citep{Bennett1980}.

\subsection{Hierarchical Hypergames}

Hypergame theory introduces a framework for modelling individual agents' subjective interpretations of the conflict they are involved in. By incorporating higher-order perceptions~\citep{Bennett1980}, hypergames enable the formalisation of recursive reasoning -- e.g., a decision-maker $r$ might consider what $q$ believes about how $p$ perceives the situation. These are called $m$-th order expectations, where $m$ corresponds to the depth of the perceptual chain. ~\citep{Wang1988} later formalised and extended Bennet's framework into a hierarchical hypergame analysis structure, addressing the overly abstract nature of earlier definitions and proposing a more rigorous, standardised representation of multi-level perceptual games. In this extended formalism, a $0^{th}$-level hypergame is equivalent to a standard game-theoretic without misaligned perceptions and is defined as a tuple $H^0=(N,P,S,U,V)$, where:
\begin{itemize}
    \item $N = \{i \mid 1 \leq i \leq n\}$ is the set of players;
    \item $P = \{P_i \mid i \in N\}$ is the set of players option sets, where:
    \begin{itemize}
        \item $P_i$ is the set of actions available to player $i$;
    \end{itemize}
    \item $S = \{S_i \mid i \in N\}$ is the set of players' strategy sets, where:
    \begin{itemize}
        \item $S_i \subseteq P_i$ is the set of actions player $i$ considers for $G$;
    \end{itemize}
    \item $U$ is the set of all possible outcomes, where:
    \begin{itemize}
        \item Each outcome $u \in U$ is a tuple of chosen strategies $u = (s_1,\ldots,s_n)$;
        \item $U = \bigtimes_{i \in N} S_i$ represents all possible strategy combinations;
    \end{itemize}
    \item $V = \{V_i \mid i \in N\}$ is the set of player preferences, where for each player $i$:
    \begin{itemize}
        \item $V_i = (U, R_i)$ where $R_i$ is a preference relation over $U$, or
        \item $V_i = (U, u_i)$ where $u_i: U \rightarrow \mathbb{R}$ is a utility function.
    \end{itemize}
\end{itemize}

The distinction between $P_i$ and $S_i$ allows the modelling of agents' beliefs about physical or cognitive constraints. A zero-level hypergame assumes shared perceptions -- there are no subjective differences between agents' views of the game.

\paragraph{Higher-order Reasoning}

To support higher-order reasoning, Wang et al. introduced the notion of \textit{perceptual functions}. Each player's \textit{viewpoint} encodes their holistic interpretation of the game, while individual perceptions capture their interpretation of specific components. For any player $i$, a perception is defined as a mapping:
\begin{equation*}
    f_i:\Gamma_i \to \Gamma_{ij},
\end{equation*}

where:
\begin{itemize}
    \item $\Gamma_j$ is a component from player $j$'s game (from $\{N,P,S,U,V\}$);
    \item $\Gamma_{ij}$ denotes player $i$'s internal model of $\Gamma_j$;
    \item For each element $\gamma \in \Gamma_j$, $\varphi = f_i(\gamma)$ is $i$'s perceived \textit{image} of $\gamma$.
\end{itemize}

In the case of nested beliefs, player $ i$'s interpretation is composed via the product mapping:
\begin{equation*}
    f_i \circ f_j : \Gamma_k \rightarrow \Gamma_{ijk}.
\end{equation*}
The resulting image is the product of individual perceptual functions:
\begin{equation*}
    \varphi= f_i(f_j(\gamma)) = f_i \circ f_j(\gamma) = f_{ij}(\gamma).    
\end{equation*}

Misperceptions can arise across all game components when the image $\varphi=f_i(\gamma)$ deviates from the true $\gamma \in \Gamma_j$. A perception is:
\begin{itemize}
    \item Underperceived if it omits elements present in $\Gamma_j$;
    \item Overperceived if it includes elements not present in $\Gamma_j$;
    \item Otherwise, a misperception if $\varphi \neq \gamma$.
\end{itemize}

\paragraph{Higher-order Games}

In cases where players develop misaligned perceptions and expectations about others' games, we define higher-order hypergames as follows:
\begin{itemize}
    \item A \textbf{first-level hypergame} assumes that players are unaware of any perceptual misalignment. Each player $i$' viewpoint is captured by their own perceptual game $G_i$, forming the composite hypergame $H^1=\{G_i \mid 1 \leq i \leq n\}$. Each perceptual game is defined using potentially distinct sets $N_i,P_i,S_i,U_i,V_i$.
    \item A \textbf{second-level hypergame} arises when at least one player is aware that others may have different viewpoints. $H^2=\{H_{i}^1 \mid 1 \leq i \leq n\}$, where each $H^1_i$ now contains the set of perceptual games $G_{ij}=(N_{ij},P_{ij},S_{ij},U_{ij},V_{ij})$ that player $i$ believes player $j$ uses: $H_i^1 = \{G_{ij} \mid 1 \leq j \leq m\}$.
    \item A \textbf{third-level hypergame} extends this recursion once more: a player $i$ models what player $j$ believes about what player $k$' game. We have $H^3=\{H_{i}^2 \mid 1 \leq i \leq n\}$, where each individual game is composed of first-level hypergames $H_{i}^2=\{H_{ij}^1 \mid 1 \leq j \leq m\}$ and each first-level hypergame is a set of standard games $H_{ij}^1=\{G_{ijk} \mid 1 \leq k \leq h\}$, where each percpetual game is $G_{ijk}=(N_{ijk},P_{ijk},S_{ijk},U_{ijk},V_{ijk})$.
\end{itemize}

At each level, the formal structure is extended by an additional layer of subjective (hyper)games built on nested perceptions. While the mathematical framework supports $L^{th}$-order hypergames of arbitrary depth, we restrict the discussion to level 3, reflecting empirical observations that human strategic reasoning seldom exceeds three levels~\citep{camerer2004cognitivehieararchy} and balancing expressiveness and computational tractability.

\paragraph{Hypergame Nash Equilibrium}

The Hypergame Nash Equilibrium (HNE) extends the classical Nash Equilibrium to settings where players may have divergent perceptions of the game~\citep{Wang1988}. The HNE formalizss mutual best responses in this context of misaligned perceptions; that is, the HNE is an NE in each player's corresponding subjective game. Given a strategy profile $a^*=(a^*_i,a^*_{-i})$:
\begin{itemize}
    \item In a first-level hypergame $H^1=\{G_i \mid 1 \leq i \leq n\}$, $a^*$ is a HNE if, for each player $i \in N$, $a^*$ is a NE in their subjective game $G_i$;
    \item In a second-level hypergame $H^2=\{H^1_i \mid 1 \leq i \leq n\}$, $a^*$ is a HNE if, for each player $i\in N$ and $j \in N_i$, $a^*$ is a NE in $G_{ij}$;
    \item In a third-level hypergame $H^3=\{H_{i}^2 \mid 1 \leq i \leq n\}$, $a^*$ is a HNE if, for each player $i\in N$, $j \in N_i$ and $k \in N_{ij}$, $a^*$ is a NE in $G_{ijk}$.
\end{itemize}

\subsection{Hypergame Normal Form}

The normal form is a foundational representation in game theory used to express strategic interactions among decision-makers, in which players' actions and the corresponding payoffs are captured in a table, typically referred to as the payoff matrix~\citep{Rasmusen1990-RASGAI}. The Hypergame Normal Form (HNF) extends the standard normal form to capture a single player's beliefs -- drawing on decision-theoretic concepts~\citep{Hansson2011} -- about their opponent's reasoning~\citep{vane2000thesis}. Originally proposed for analysing complex military planning scenarios, HNF enables an agent to reason over multiple possible opponent strategies, each corresponding to a different perceptual subgame. Table~\ref{tab:hnfrps} illustrates an HNF matrix for the rock-paper-scissors game used in previous examples from the row player’s perspective. The HNF consists of:
\begin{itemize}
    \item The \textbf{Full Game} (bottom right quadrant) represents the ``expert's view'' -- a standard $n\times m$ game-theoretic payoff matrix, where $n$ and $m$ denote the set of actions available to the row and column players, respectively;
    \item The \textbf{Column Mixed Strategies} (top right quadrant) or CMS represent possible opponent strategies, including:
    \begin{itemize}
        \item $\mathbf{C_0}$: the Nash equilibrium mixed strategy (NEMS);
        \item $\mathbf{(C_1, \ldots, C_k)}$: the row player’s belief-induced subgames representing alternative opponent mixed strategies over $(S_1,\ldots,S_m)$;
        \item $\mathbf{C_\Sigma}$: a weighted summary of beliefs, computed as:
            \[S_j=\sum_{k=0}^{K-1} P_kc_{kj},\quad \text{for } j = 0, 1, \ldots, n\]
            where $P_k$ is the weight for belief context $k$ and $c_{kj}$ is the weight for strategy $j$ in context $k$.
    \end{itemize}
    \item The \textbf{Row's Belief Context} (upper left quadrant) contains probabilities assigned to each opponent CMS. These are analogous to perceptual functions in multi-level hypergames but focused on strategy selection rather than game structure. They capture the player’s beliefs about the likelihood of each opponent's reasoning mode, that is, the probability that the corresponding subgame (CMS row) is active;
    \item The \textbf{Row Mixed Strategy} (RMS) zone (bottom left quadrant) captures the row player’s responses to each belief context, resembling second-order beliefs in multi-level hypergames. RMS entries are not prescriptive but inform planning.
\end{itemize}

\begin{table}[h]
  \centering
  \renewcommand{\arraystretch}{1.5}
  \begin{tabular}{cccc>{\columncolor[gray]{0.8}}ccccc} 
     & & & & &\multicolumn{3}{c}{\bfseries Column Mixed Strategy} \\
     \multicolumn{4}{c}{\bfseries Row's Belief Context}&\bm{$C_\Sigma$}&\bm{$S_1$}&\bm{$S_2$}&\bm{$S_3$} \\
     $P_2$ & 0.5 & & & \bm{$C_2$} & $\frac{1}{2}$ & $\frac{1}{2}$ & --- \\
     $P_1$ & & 0.25 & & \bm{$C_1$} & 0.326 & 0.3456 & 0.3284 \\
     $P_0$ & & & 0.25 & \bm{$C_0$} & $\frac{1}{3}$ & $\frac{1}{3}$ & $\frac{1}{3}$ \\
     \rowcolor{lightgray} \bm{$H_1$} & \bm{$R_2$} & \bm{$R_1$} & \bm{$R_0$} & & \bfseries Rock & \bfseries Paper & \bfseries Scissors \\
     $H_{11}$ & 0.2 & 0.3284 & $\frac{1}{3}$ & \bfseries Rock & 0 & -1 & 1 \\
     $H_{12}$ & 0.6 & 0.326 & $\frac{1}{3}$ &\bfseries Paper & 1 & 0 & -1 \\
     $H_{13}$ & 0.2 & 0.3456 & $\frac{1}{3}$ &\bfseries Scissors & -1 & 1 & 0 \\
     \multicolumn{4}{c}{\bfseries Row Mixed Strategy} & & \multicolumn{3}{c}{\bfseries Full Game}
  \end{tabular}
  \caption{Hypergame Normal Form matrix for rock-paper-scissors comprising of: (1) Row's Belief Context with probabilities $P_k$ for each belief context, (2) Column Mixed Strategies showing opponent mixed strategies $C_k$ and their aggregate $C_\Sigma$, (3) Row Mixed Strategy containing $R_k$ responses to each belief context, and (4) Full Game with cardinal utilities. Hyperstrategies $H_{ij}$ represent evaluated plans, expressed as probability distributions over row strategies. Each $C_k$ and $R_k$ represents a mixed strategy vector summing to 1.}
  \label{tab:hnfrps}
\end{table}

The game in Table~\ref{tab:hnfrps} defines the row player's beliefs about the column player in rock-paper-scissors. $C_0$ corresponds to the mixed strategy Nash equilibrium (uniform randomisation), while $C_1$ reflects empirical distributions derived from online gameplay data~\citep{batzilis19rockpaperscissors}. In $C_2$, the row player believes that the opponent cannot play Scissors. The belief context assigns the highest probability to $C_2$, indicating that the row player expects this restricted subgame to be most likely. The RMS entries mirror responses to each $C_k$. Additionally, the HNF matrix can include expected utility (EU) and hypergame expected utility (HEU) calculations as additional rows beneath the matrix -- omitted in Table~\ref{tab:hnfrps} for clarity.

\paragraph{Hyperstrategies}
Hyperstrategies are mixed strategies over RMSs, represented as vectors of assigned probabilities adjacent to the RMS quadrant. We include a hypothetical hyperstrategy $H_1$ in Table~\ref{tab:hnfrps} for illustration. Vane~\citep{vane2000thesis} defines four hyperstrategies that meet the following efficiency criterion:
\begin{equation*}
    \text{Eff}(hs) \leftrightarrow \text{EU}(hs,C_\Sigma) \geq \text{EU}(\text{NEMS}(m \times n)).
\end{equation*}

\noindent These are defined as follows:

\begin{itemize}
    \item \textbf{NEMS}: NEMS is the baseline Nash equilibrium mixed strategy derived from the full game;
    \item \textbf{Modelling Opponent (MO)}: MO is a decision-theoretic strategy selecting the best response based on CMS and RMS utilities;
    \item \textbf{Pick Subgame (PS)}: PS selects the Nash equilibrium of the subgame identified by MO;
    \item \textbf{Weighted Subgame (WS)}: WS is a hybrid strategy that chooses the best option based on the combination of the MO-specified CMS and $R_0$ weighted by the corresponding belief context probabilities. 
\end{itemize}

\paragraph{Hypergame Expected Utility}
Vane~\citep{vane2000dtgt} introduced hypergame expected utility (HEU) as a bridge between decision theory and game theory. HEU incorporates a parameter $g$ -- the ``fear of being outguessed'' -- and the worst-case outcome $G$, defined as the lowest expected utility for the row player. Then, given a hyperstrategy $hs$, we define the HEU as:
\begin{equation*}
    \text{HEU}(hs,g) = (1-g)\cdot\text{EU}(hs,C_{\Sigma})+g\cdot\text{EU}(hs,G).
\end{equation*}

The relative performance of hyperstrategies such as MO, PS, and WS varies with the uncertainty parameter $g$, with NEMS becoming increasingly dominant as $g$ increases. The analysis of $g$’s effects -- detailed in~\citep{vane2000dtgt} -- suggests that when agents hold accurate beliefs about their opponent’s strategy, HEU-optimised hyperstrategies can outperform standard equilibrium analysis.

\section{Hypergame Theory in Multi-Agent Systems}
\label{sec:sysrev}
While hypergame theory is a model of bounded rationality that can provide multi-agent systems with a framework that can model agent heterogeneity and environmental uncertainties to a greater level of detail than probabilistic methods, it was originally designed to be an analytical framework targeting post-hoc studies of complex conflicts instead of providing tools for the decision-maker on the spot. While hypergame models have yet to become prevalent in dynamic settings that multi-agent systems entail, they have the potential to overcome the constraints posed by typical game-theoretic models on the complexity that agent-based frameworks can capture~\citep{Sasaki2021multiagaentdecisionsystems}. In this section, motivated by this identified potential, we provide a systematic review of hypergame-theoretic works focusing on studies that entail agent-based applications or the potential for integration into dynamic agentic systems. 

\subsection{Methodology}

\begin{figure}[htbp]
\centering
\includegraphics[width=\linewidth]{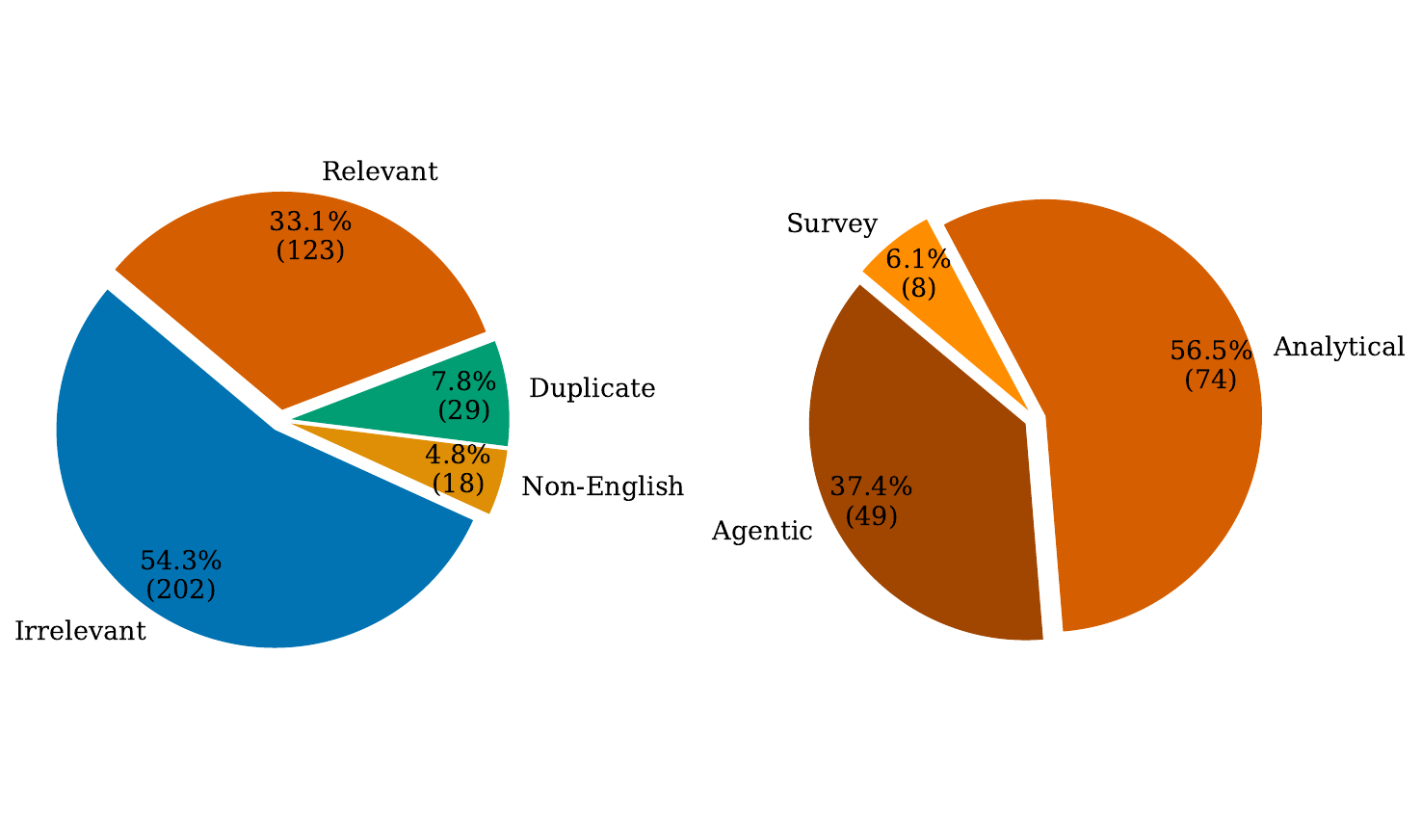}
\caption{Distribution of the evaluated 380 papers based on our selection criteria. The left chart illustrates the composition of papers eliminated due to irrelevance, duplication, or being written in a language other than English. The chart to the right illustrates the agent-compatibility evaluation that resulted in the final 49 papers considered for our review.}
\label{fig:selection}
\end{figure}

Our systematic survey of hypergame theory in the context of multi-agent systems follows a multi-step screening and classification process. Our goal was to identify works in which hypergame-theoretic models are not only referenced but explicitly applied to support agentic reasoning, decision-making, or strategic interaction within MAS-compatible contexts. Figure~\ref{fig:selection} provides an overview of this selection funnel. Starting with an inclusive set of keyword-matched results, we progressively filtered out duplicates, non-English works, irrelevant uses, and purely analytical studies lacking actionable agent-level integration. The steps are outlined below:

\paragraph{Initial Collection}

We conducted a keyword-based query based on Google Scholar using the term ``hypergame theory”, which yielded 380 unique results as of April 2026.

\paragraph{Duplicate Removal}
We removed 29 entries due to duplication. Removed entries included different versions of the same manuscript (e.g., preprints and published versions) as well as overlapping content such as thesis chapters and standalone publications.

\paragraph{Language Filter}

We excluded 18 papers for being written in languages other than English.

\paragraph{Surveys and Meta-reviews}

We identified 8 publications that were either surveys of hypergame theory or broader reviews that discussed hypergames as part of a more general framework. These were retained for comparative purposes but not included in the core set analysed for model classification.

\paragraph{Relevance Filtering}

While hypergame theory is a distinct subfield within game theory, the term is occasionally used in unrelated contexts or cited without substantive application. To ensure relevance, we manually examined each entry’s abstract, methods, references, and in-text keyword usage. We further excluded 202 papers that either lacked an actual implementation of hypergames, used the concept only metaphorically, or mentioned it in passing without integrating it into the modelling framework.

\paragraph{Agent-compatibility Filter}

From the remaining 123 works, we applied a final inclusion criterion focused on agent compatibility. Specifically, we retained only those works in which hypergames were used to support or influence agent behaviour, strategic decision-making, or reasoning under uncertainty -- either in simulated, practical, or theoretical settings with agent-level implications. This step excluded 74 additional papers that employed hypergames solely from an analyst perspective or in ways that did not feed back into the agent's own actions or perceptions.

As a result, we identified a final corpus of 49 papers featuring agent-compatible hypergame-theoretic models. Although only three of these explicitly formalise the agent architecture, the majority assume rational decision-makers with embedded utilities or subjective games. Due to inconsistencies in architectural specifications, our review focuses on the exclusion of a discussion of implicit and explicit agent designs.

\subsection{Hypergame-theoretic Publication Trends}
\begin{figure}[htbp]
\centering
\includegraphics[width=\linewidth]{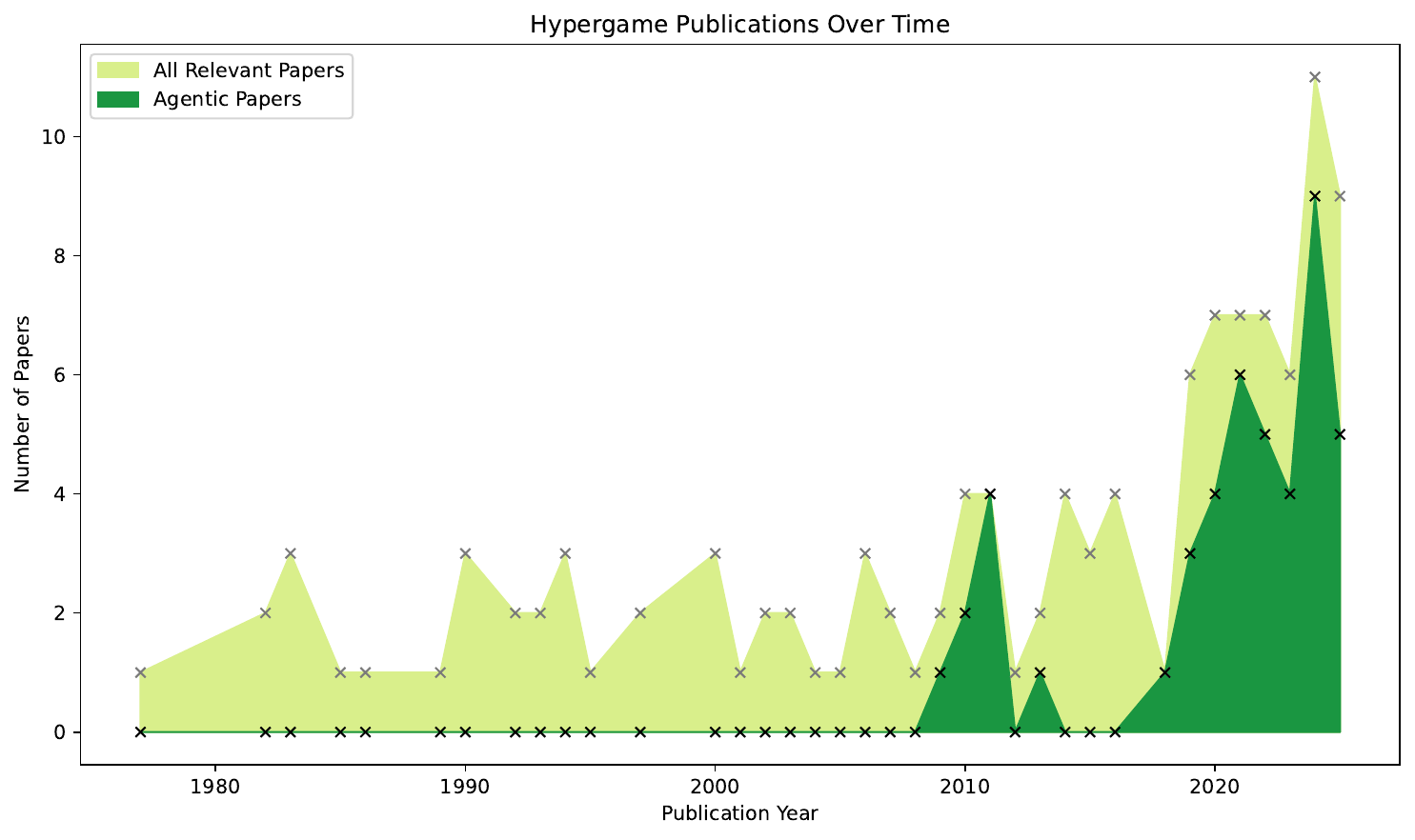}
\caption{Annual publication counts for all hypergame-relevant papers and their subset involving agentic applications. Counts are visualised up to 2025.}
\label{fig:time}
\end{figure}

Bennett introduced the theory of hypergames as an analytical framework extending standard game-theoretic frameworks, emphasising the modelling ability to account for perceptual misalignments. Vane's hypergame normal form revised Bennett's hypergame representation, providing an extended decision-theoretic matrix form focusing on one player's belief systems and reasoning. Early applications of both frameworks consisted solely of post-hoc or hypothetical analyses, with no demonstration of how a decision-maker can benefit from hypergame theory in real time. While both original authors repeatedly highlighted the potential of hypergames for dynamic agent-based applications and both attempted ``live'' hypergame analyses, the adoption of hypergame theory in the multi-agent literature, in particular, remains to take off. Nonetheless, our review reveals a trend of increasing numbers of MAS-related hypergame-theoretic publications in recent years. We see a shift towards agent-based applications, as illustrated in Figure~\ref{fig:time}.
Interestingly,~\citep{kovach2015hypergamereview} stated that ``Unfortunately very little in the way of hypergame theory application has been done in the arena of cyber warfare'', while a decade later the current shift is dominantly lead by attack-defense modeling and simulations in cybersecurity -- as demonstrated by Figures~\ref{fig:app_type_main},~\ref{fig:integ_type_main},~\ref{fig:concept-domain} and the various cybersecurity-related surveys covering hypergame-based approaches we discuss in section~\ref{sec:relwork}.\\

Hypergame theory is a niche field that has not yet gained traction, with a modest number of publications, dominated by a historical concentration of contributions from the original authors and their collaborators. To investigate whether recent trends -- such as increasing domain diversity and methodological variation -- are accompanied by a broader dissemination of authorship, we conducted an analysis of author frequencies and co-authorship patterns across the 49 papers reviewed. We extracted all named authors from each publication, accounting for different name formats and removing placeholder or malformed entries. First, we identified individual authors associated with at least three of the reviewed papers. Then, to identify collaborative groups, we constructed a co-authorship graph in which nodes represent individual authors and edges indicate co-authorship on at least one paper. Groups were then defined using connected components of this graph, with each component representing a set of authors linked by direct or indirect collaboration. For each group, we computed the number of papers coauthored by at least two members, thereby capturing the extent and productivity of distinct author clusters.

\begin{figure}[htbp]
\centering
\includegraphics[width=\linewidth]{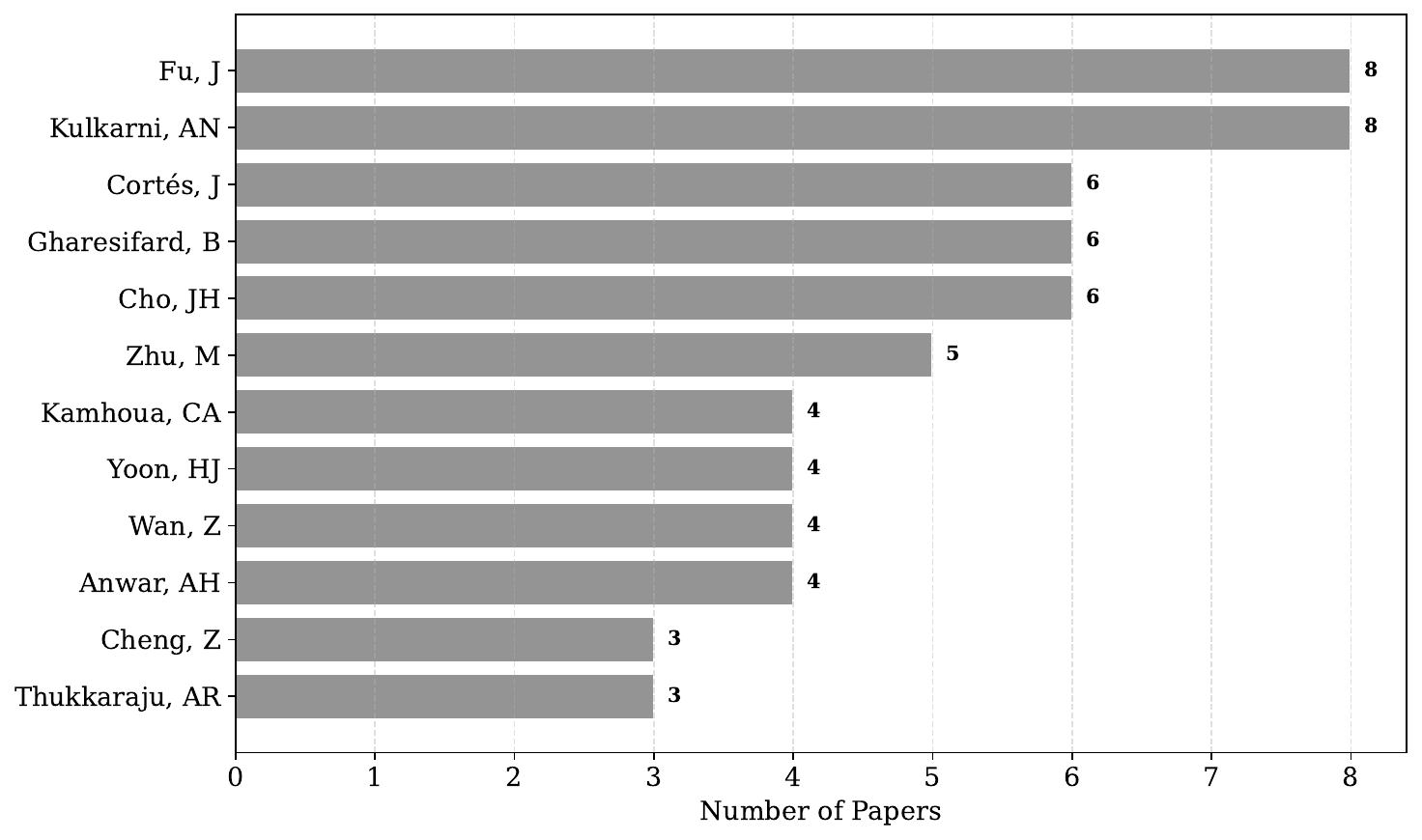}
\caption{Publication counts per authors with at least 3 associated works.}
\label{fig:authors}
\end{figure}

As shown in Figure~\ref{fig:authors}, twelve authors are associated with at least three papers, with nearly a third of reviewed papers attributable to two recurring author pairs: Kulkarni \& Fu and Gharesifard \& Cortés. Furthermore, as shown in Table~\ref{tab:author_groups}, we identified four groups of collaborators with at least three relevant publications, with Group 5 -- Kulkarni \& Fu et al. -- accounting for a quarter of publications. At the same time, nearly half of the papers originate from authors or groups with fewer than three publications, suggesting a shift from concentrated research programs toward more diverse, application-specific uses of hypergames, potentially reflecting growing interest and wider adoption of the theory across domains -- a finding well-aligned with the publication trends identified via our classification framework and Figure~\ref{fig:time}.

\begin{table}[ht]
\centering
\caption{Identified coauthor groups with at least 3 shared papers. Member names are shown by surname only.}
\begin{tabular}{ccl}
\toprule
\textbf{Group} & \textbf{Shared Papers} & \textbf{Members} \\
\midrule
G5 & 11 & \begin{tabular}[t]{@{}l@{}}Cohen, Fried, Fu, Gaglione, Han, Hemida, Kamhoua,\\ Kulkarni, Leslie, Li, Liu, Luo, Ma, Shi, Udupa, Xi\end{tabular} \\
G1 & 8  & \begin{tabular}[t]{@{}l@{}}Anwar, Cho, Kamhoua, Singh, Wan, Zhu\end{tabular} \\
G6 & 6  & Cortés, Gharesifard \\
G8 & 4  & \begin{tabular}[t]{@{}l@{}}Chen, Cheng, Gao, Guan, He, Hong, Ma, Yuan\end{tabular} \\
\bottomrule
\end{tabular}
\label{tab:author_groups}
\end{table}

\subsection{Taxonomy of Hypergame Models}

In this section, we provide a detailed taxonomy of hypergame models identified across the surveyed papers. We categorise the models according to three key dimensions: the foundational hypergame concept and game dynamics, formalisations of hypergame frameworks, and the computational tasks hypergame-theoretic approaches address. This taxonomy facilitates a structured understanding of how different hypergame implementations vary in their theoretical assumptions, representational complexity, and practical application scenarios, enabling researchers to select or develop hypergame models tailored to specific multi-agent system requirements.

\paragraph{Game Dynamics}

First, we looked at whether the hypergame models entail a single round of interaction or repeated, iterative decision-making. The distribution across hypergame concepts and game dynamics is characterised by 33 applications that entail repeated interactions, while 16 models rely on one-shot games. This suggests a shift in application trends, as historically, work has primarily involved one-shot post-hoc studies.

\paragraph{Hypergame Concept}

We also distinguished works by whether they were inspired by the original~\citep{Bennett1980} or extended~\citep{Wang1988} hypergame framework -- collectively denoted as \textbf{Multi-level Hypergames (MLH)} -- or based on the \textbf{Hypergame Normal Form (HNF)}. We found 11 papers that refer to HNF, 36 rely on Multi-level Hypergames, and 2 papers use components from both concepts, which we denote as \textbf{Hybrid} approaches.

\paragraph{Hypergame Formalizations}

Next, we identified common practical hypergame model definitions, which we classify as follows:

\begin{itemize}
    \item \textbf{HNF-based} solutions populate complete Hypergame Normal Form (HNF) matrices, which agents directly utilise for decision-making.
    \item \textbf{HEU-based} methods utilise the Hypergame Expected Utility derived from HNF to help agents select optimal strategies.
    \item \textbf{L-th Order Multi-level Hypergames} populate hierarchical hypergame structures, enabling agents to form nested beliefs.
    \item \textbf{Flattened L-th Order Multi-level Hypergames} employ subjective games but exclude the hierarchical tree structure.
    \item \textbf{Bayesian L-th Order Hypergames} implement hybrid belief models in which hypergames represent perceptions, but agents perform decision-making using Bayesian reasoning.
    \item \textbf{Perceptual Game-based} implementations enable agents to develop hypergame-like beliefs without defining a complete multi-level hypergame structure.
    \item \textbf{HNE-based} frameworks involve decision-makers explicitly relying on the Hypergame Nash Equilibrium concept.
    \item \textbf{Graph-based} methods incorporate spatial hypergames and strategies utilising graph-based representations.
    \item \textbf{Stackelberg Hypergames} extend the leader-follower mechanism of Stackelberg games by integrating hypergame-theoretic subjective perceptions.
\end{itemize}

\begin{figure}[htbp]
\centering
\includegraphics[width=\linewidth]{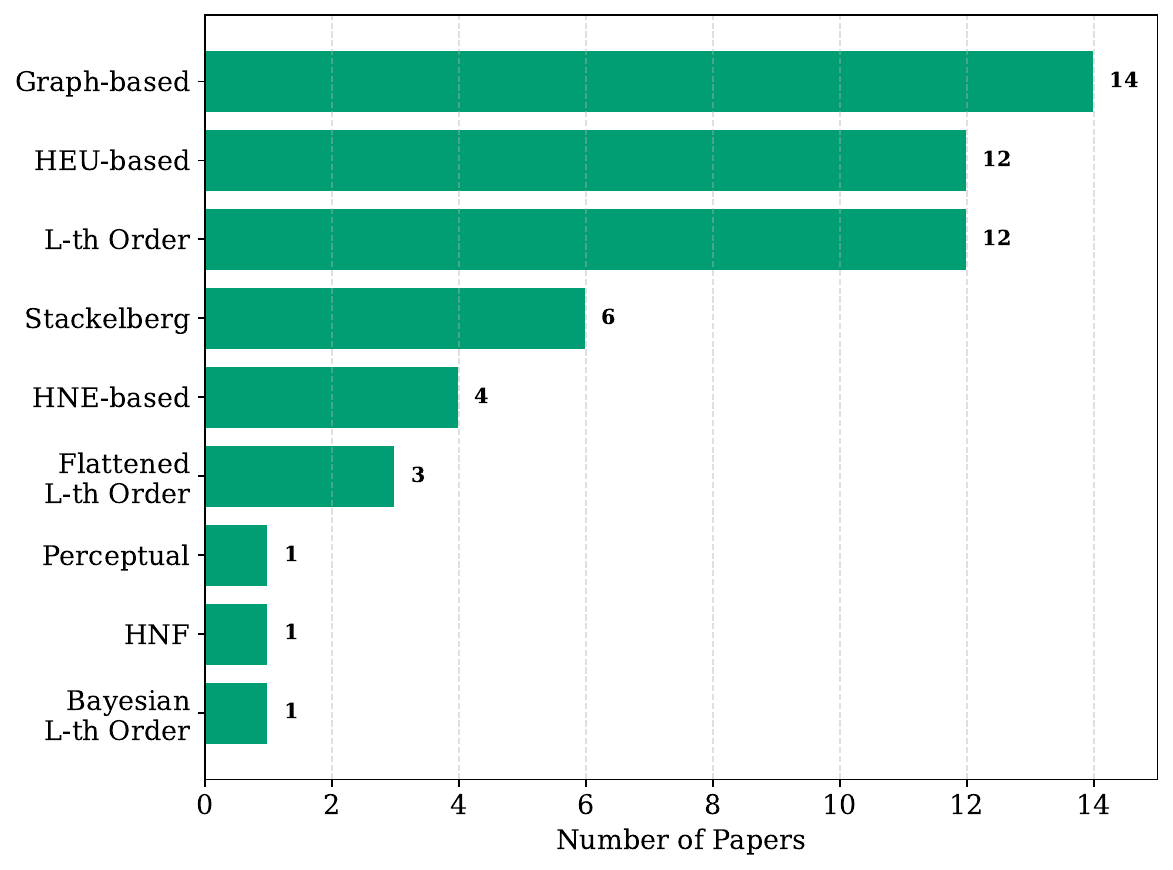}
\caption{Distribution of model types. Hybrid models that span multiple categories are counted in each relevant type.}
\label{fig:tax_models}
\end{figure}

The distribution of model usage across the surveyed papers is depicted in Figure~\ref{fig:tax_models}: the majority of works implement a model based on the typical multi-level hypergame concept (L-th Order, Flattened L-th Order and Bayesian L-th Order) or hypergames on graphs. 5 works entail hybrid solutions, where, in addition to populating a hypergame model, a hypergame-theoretic solution concept (HEU or HNE) is also implemented. In two cases, the hybrid approach entailed cross-framework implementations: multi-level-hypergame-based models integrating the hypergame expected utility concept from HNF.

\subsection{Hypergames in Practice}

Hypergames, despite being originally developed as a post-hoc analytical framework, provide a promising tool for modeling realistic interaction scenarios with misaligned perceptions in multi-agent systems. Practical implementations employing, to varying extents, hypergame concepts emerge in such diverse fields as cybersecurity, robotics, and social simulations. To identify current patterns in the practical implementation of hypergames, as well as the challenges and open research problems, we analyze the papers that implement hypergame concepts in terms of the application domain (Section~\ref{sec:application}), hypergame integration methodology (Section~\ref{sec:int-methodology}), integration fidelity (Section~\ref{sec:fidelity}), and finally, the role of the hypergame model within the given framework (Section~\ref{sec:task}). 

\subsubsection{Application Domains}
\label{sec:application}

\begin{figure}[htbp]
\centering
\includegraphics[width=\linewidth]{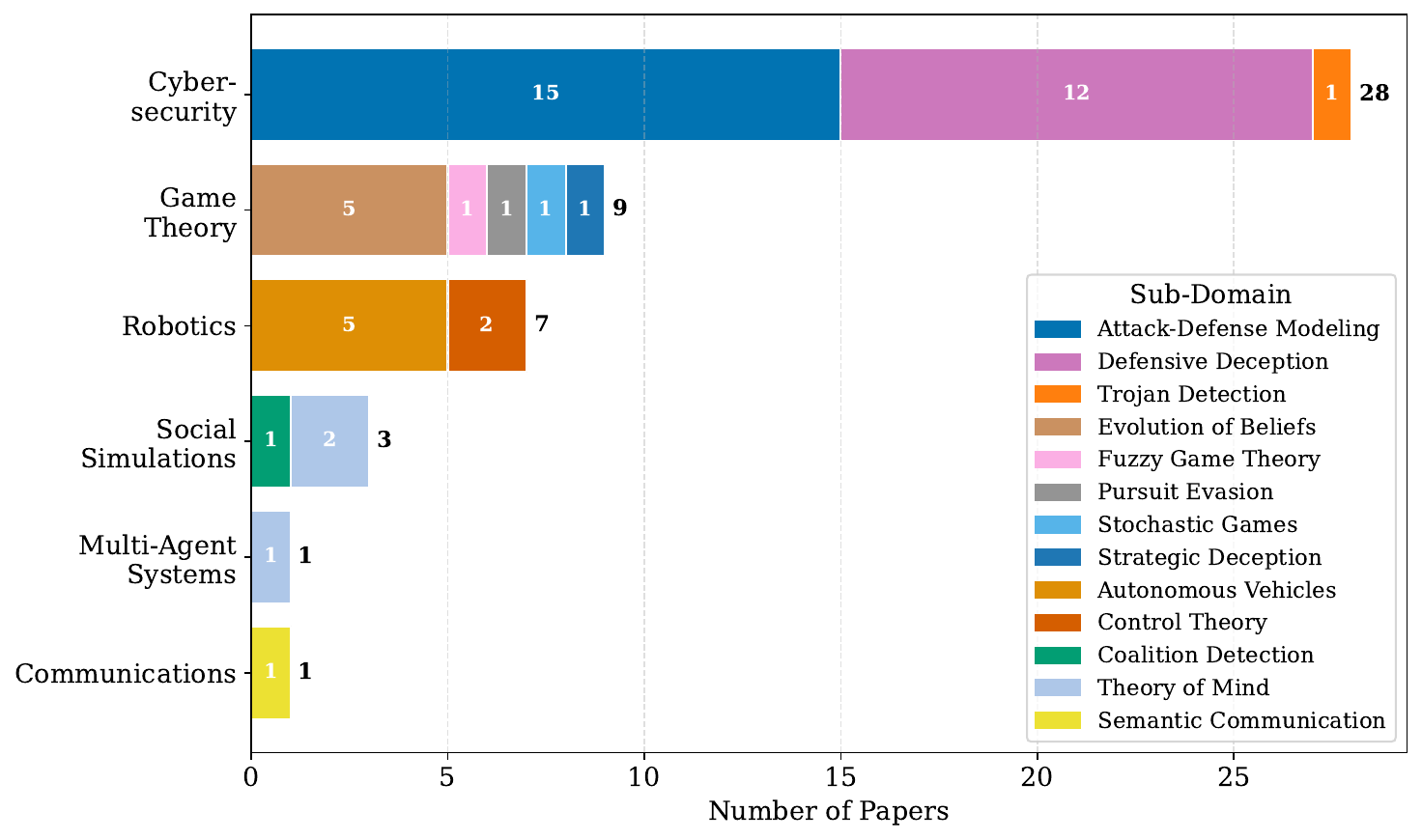}
\caption{Distribution of agent-compatible hypergame papers across high-level application domains and their associated sub-domains, where sub-domains can overlap between application domains.}
\label{fig:domain}
\end{figure}

To investigate how hypergame theory has been adopted across different areas of research and application, we classified the 49 agent-compatible papers according to their primary domain of use. Each paper was assigned to one of six high-level domains, which reflect the thematic focus of the study: \textit{Game Theory}, \textit{Cybersecurity}, \textit{Social Simulations}, \textit{Robotics}, \textit{Multi-agent Systems}, and \textit{Communications}. Within each domain, we identified recurring sub-domains that capture specific use cases, each briefly introduced in the following subsections and highlighted in \textbf{bold} to aid navigation. In Figure~\ref{fig:domain}, we provide an overview of the reviewed works across the identified domains and sub-domains, which are summarized as follows.\\

\noindent\textbf{Game Theory} Works that advance game theory in general are categorised under the general domain of \textit{Game Theory}. These papers utilise hypergame-theoretic tools to define equilibrium concepts, model learning dynamics under misperception, or devise strategic responses to asymmetry and deception.
For example, ~\citep{song2009fuzzyhypergame} integrates \textbf{Fuzzy Game Theory} with hypergames by incorporating fuzzy strategies and imprecise preference perceptions. In contrast, ~\citep{yuan2024equilibrium} focuses on \textbf{Deception} and uses hypergames to analyse how misinformation affects equilibrium strategies. Additionally, ~\citep{li2024hypergameQlearning} proposes a hypergame-based Q-learning approach to solve \textbf{Pursuit-Evasion} games, while ~\citep{chongyang2023stochasticgames} develops a hypergame-theoretic model of deception in concurrent \textbf{Stochastic Games}. Lastly, various works leverage the hypergame-theoretic \textbf{Evolution of Beliefs} in adversarial games characterised by misperceptions. These studies define mechanisms for belief updates~\citep{gharesifard2010evolution, gharesifard2012evolution} that target the learning of equilibria~\citep{gharesifard2011exploration, gharesifard2011learningequilibria} and belief manipulation~\citep{gharesifard2011stealthy}.\\

\noindent\textbf{Cybersecurity} Cybersecurity represents the most active domain for agent-compatible hypergame applications. This prevalence can be attributed to the inherent suitability of hypergames for capturing misinformation and modelling active deception -- both critical issues within cybersecurity contexts. The majority of the surveyed works address \textbf{Attack-Defense Modeling} in general, formulating hypergames on graphs and applying linear temporal logic to guide deceptive strategy synthesis~\citep{kulkarni2021reachabilitygames,kulkarni2021hypergamegraphs,kulkarni2021labeling,lening2023dynamichypergames}. Other examples include belief manipulation~\citep{gharesifard2014stealthydeception,bakker2020infomanipulation}, mission impact assessments employing hypergame-theoretic frameworks~\citep{thukkaraju2023mia,yoon2025mia,yoon2026imia,yoon2026cyber}, and other specialized hypergame-based methodologies for attack-defense modeling~\citep{house2010hypergame,guiterrez2018,bakker2021metagames,cheng2022stackelberg}. In addition, several works target a narrower area of attack-defence modelling. \textbf{Defensive Deception} entails the use of deceptive strategies to mislead the attacker, often focusing on optimizing strategic honeypot, decoy, and false target assignment~\citep{bowei2020iobt,kulkarni2020decoyallocation,wan2021foureyedefensivedeceptionbased,anwar2020,kulkarni2024integratedresourceallocationstrategy,shen2025iot,wan2026cyber,ma2025icloud}. Alternative approaches include using defensive deception strategies to increase the cost of attacks through modeling uncertainty via hypergames~\citep{jia2024defense} and maximizing strategic surprise via sensor revealing strategies~\citep{udupa2024reactivesynthesissensorrevealing}, or exploring two-way misinformation in deception games~\citep{Cho2019deceptiongames} and defensive deception against multiple attackers~\citep{wan2023APT}.\\

\noindent\textbf{Social Simulations} Social simulations involve multi-agent-based modeling of strategic interactions within conceptualized social environments. The framework proposed by \citep{trencsenyi2025approximatinghumanstrategicreasoning} employs a hypergame-based \textbf{Theory of Mind} reasoning framework. This approach has been utilised by \citep{trencsenyi2025influencehumaninspiredagenticsophistication} to explore the ability of agentic large language models (LLMs) to engage in human-like reasoning during guessing games. Furthermore, \citep{kulkarni2025coalitiondetection} leverages hypergame theory to model \textbf{Coalition Detection} in natural language reasoning tasks performed by LLMs.\\

\noindent\textbf{Robotics} In the realm of robotic \textbf{Control Theory}, hypergame applications, for example, focus on modelling obstructed perceptions in traffic scenarios using subjective games \citep{kahn2022traffic} or uncertainty-informed trajectory planning in mixed human-robot traffic~\citep{chen2025hypergame}. In contrast, \citep{kulkarni2019reactivegames} aim to develop hypergame-based strategies in asymmetrical information environments between robots and their surroundings. Other studies in robotics explore the implementation of hypergames in the context of \textbf{Autonomous Vehicles}, particularly in the design of deceptive defence strategies for unmanned ground vehicles that operate under asymmetric perception conditions \citep{lv2024synthesisforUAV}. Others address the cognitive alignment of unmanned aerial vehicles during defensive \citep{he2022leader, hai2023uavcommunication} or exploratory \citep{dharmadhikari2021dronesexploring} missions.\\

\noindent\textbf{Multi-agent Systems} We categorise applications related to general issues within MAS under this domain.~\citep{aitchison2021hiddenrolegames} defines a set of hidden-role games targeting deceptive objectives and equipping agents with second-order \textbf{Theory of Mind} conceptualised as second-order hypergames. This framework allows agents to act on misaligned perceptions of their roles and objectives, thereby facilitating the emergence of asymmetric beliefs and belief manipulation in partially observable, mixed cooperative-competitive settings.\\

\noindent\textbf{Communications} Within the communications sector, we examine decentralised multi-user wireless networks composed of transmitters and receivers whose interactions are limited by bandwidth, channel quality, and computational resources. \textbf{Semantic communication} described a paradigm where only the most task-relevant information is transmitted while receivers utilise reasoning and reconstruction techniques to infer any missing data. In this context, \citep{thomas2024hypergametheorydecentralizedresource} applies hypergame theory to explicitly model and manage perceptual discrepancies among agents. 

\subsubsection{Hypergame Integration Methodology}
\label{sec:int-methodology}

During the paper selection process, we have already selected a subset of papers that meet the agent-compatibility criteria. We then further classified these works based on the practicality of their proposed implementation, as follows.

\begin{itemize}
    \item \textbf{Theoretical} works that extend or use hypergame theory in an application domain relying on agency. While these works do not entail a runnable system or dynamic experiments, an applicable algorithmic description and/or a practical case is provided.
    \item \textbf{Experimental} works entail hypergame-based applications that implement an experimental setting and use empirical evidence to prove or evaluate the effects of hypergame-based approaches.
    \item \textbf{Practical} applications encapsulate works that implement hypergame frameworks that are deployable and are tested in realistic simulations and real-life inspired settings.
\end{itemize}

\begin{figure}[htbp]
    \centering
    \includegraphics[width=\linewidth]{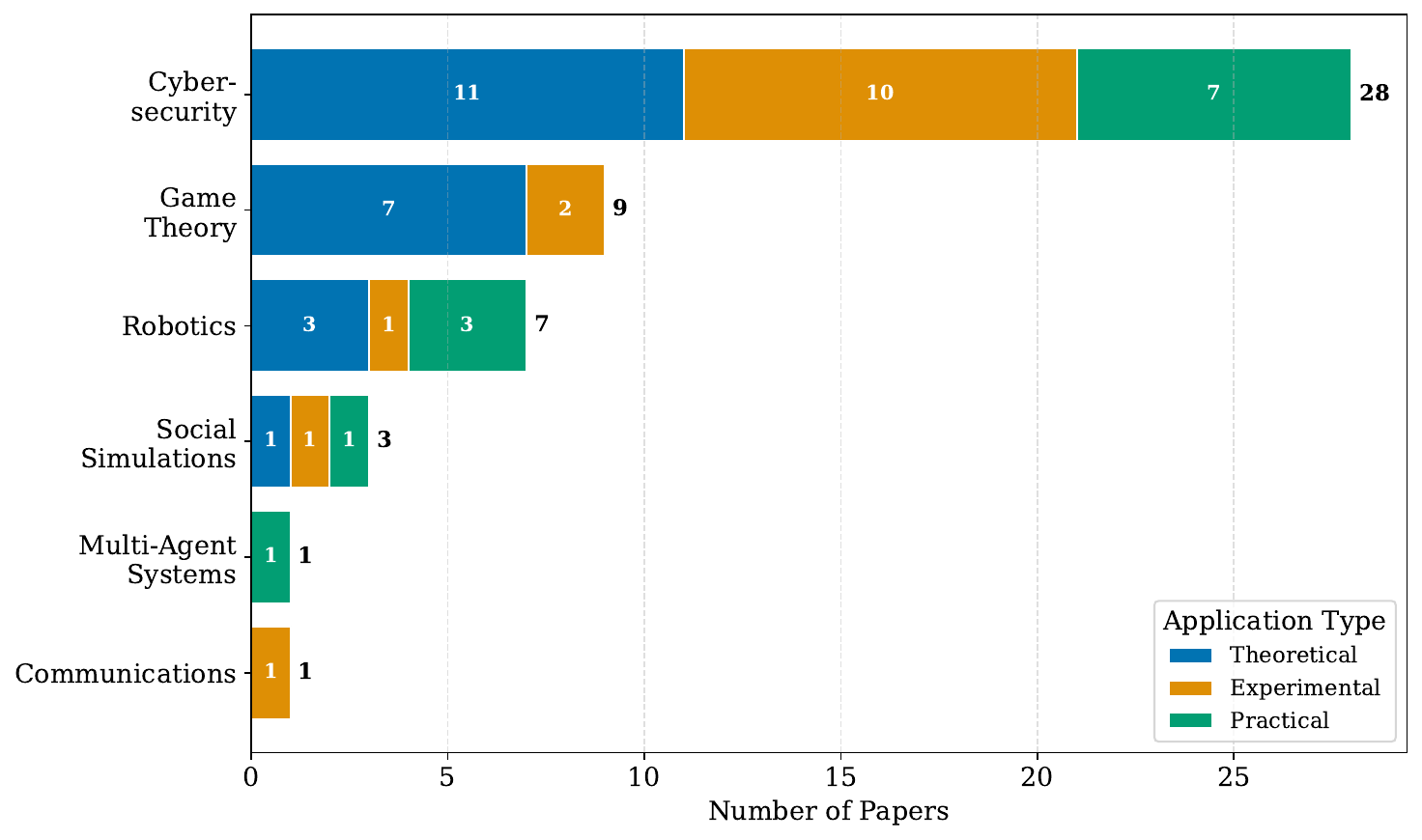}
    \caption{Domain-specific distribution of agent-compatible hypergame papers by application type. \textbf{Theoretical} works are conceptually grounded and contain algorithmic descriptions or formalized case examples, but lack implementation or simulation. \textbf{Experimental} works implement hypergame-based models in controlled settings to evaluate their effects empirically. \textbf{Practical} applications deploy fully integrated hypergame frameworks in realistic simulations or real-world-inspired scenarios.}
    \label{fig:app_type_main}
\end{figure}

Figure~\ref{fig:app_type_main} reveals that despite passing our agent-compatibility criteria, nearly half of the relevant publications fall in the theoretical category. In contrast, experimental and practical works account for 12 and 15 publications, respectively. Even across intuitively more practical domains -- such as cybersecurity, robotics, or social simulations -- theoretical papers prevail. This significant focus on the theoretical application of hypergames is well-aligned with the original design of hypergame theory as an analytical framework and with the challenges we articulated previously.

\subsubsection{Hypergame Integration Fidelity}
\label{sec:fidelity}

Next, we distinguish works based on the extent of their hypergame integration as follows:
\begin{itemize}
    \item \textbf{Conceptual} applications refer to hypergames and are inspired by hypergame-theoretic concepts but do not incorporate an explicit hypergame model.
    \item \textbf{Partial} applications are inspired by hypergame-theoretic concepts but only formally integrate components -- such as solution concepts or subjective perceptions -- instead of instantiating the entire framework;
    \item \textbf{Complete} applications entail a full hypergame model integration.
\end{itemize}

\begin{figure}[htbp]
    \centering
    \includegraphics[width=\linewidth]{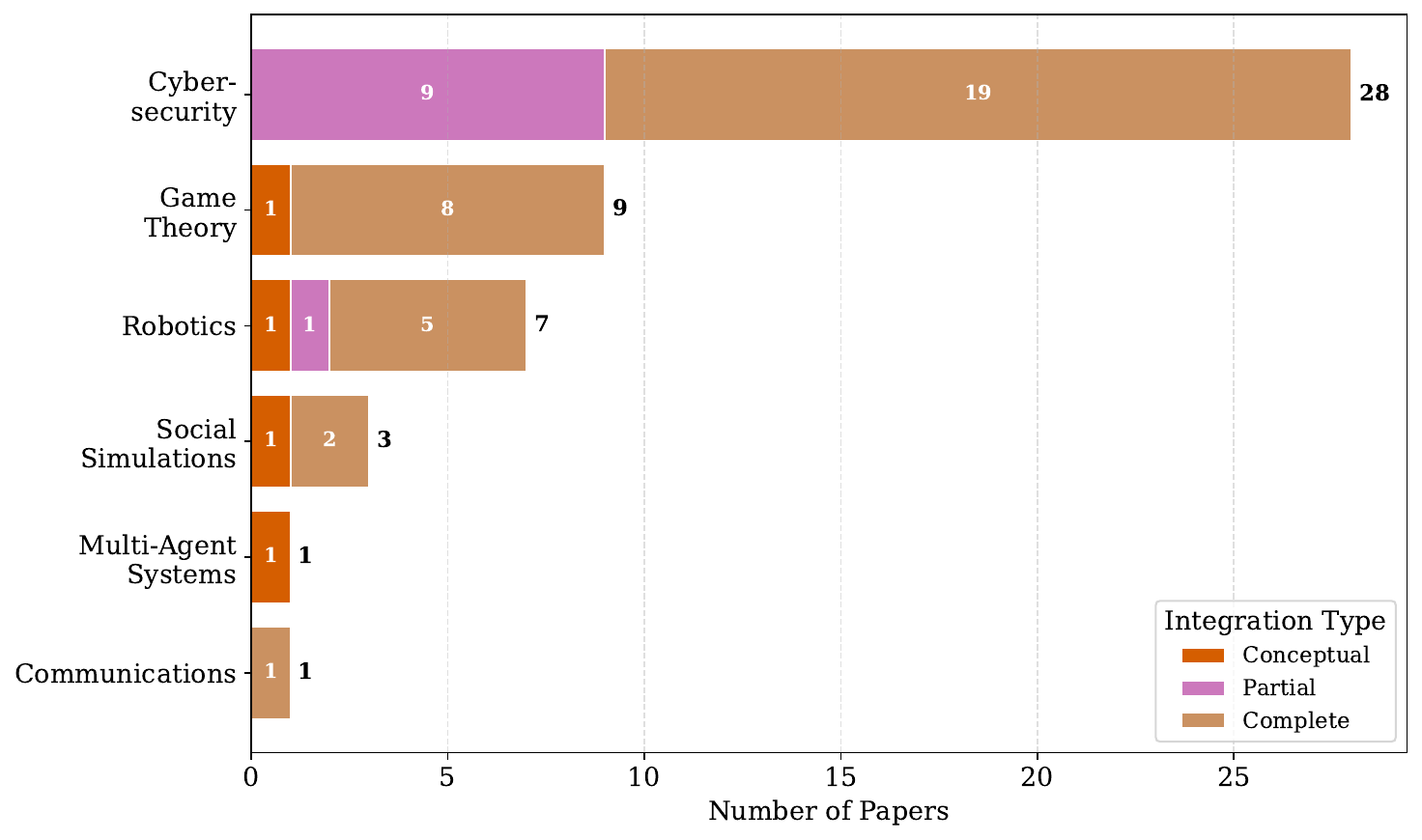}
    \caption{Domain-specific distribution of agent-compatible hypergame papers based on the extent of model integration. \textbf{Conceptual} applications reference hypergame-theoretic ideas but do not instantiate explicit models. \textbf{Partial} applications formally incorporate some hypergame components such as subjective perceptions or equilibrium concepts, without realizing a complete framework. \textbf{Complete} applications implement fully specified hypergame models within the agentic context.}
    \label{fig:integ_type_main}
\end{figure}

As depicted in Figure~\ref{fig:integ_type_main}, 35 of the surveyed papers entail complete integration of hypergame-theoretic models, 10 rely on partial implementations, and 3 approaches implement hypergames only at the conceptual level. Furthermore, the ``complete'' category is dominant across most individual domains, suggesting that, while operationalising complex hypergame analytical frameworks is non-trivial, agent-based applications benefit from high-fidelity instantiations of hypergame-theoretic models.

\subsubsection{Hypergame Task}
\label{sec:task}

\begin{figure}[htbp]
\centering
\includegraphics[width=\linewidth]{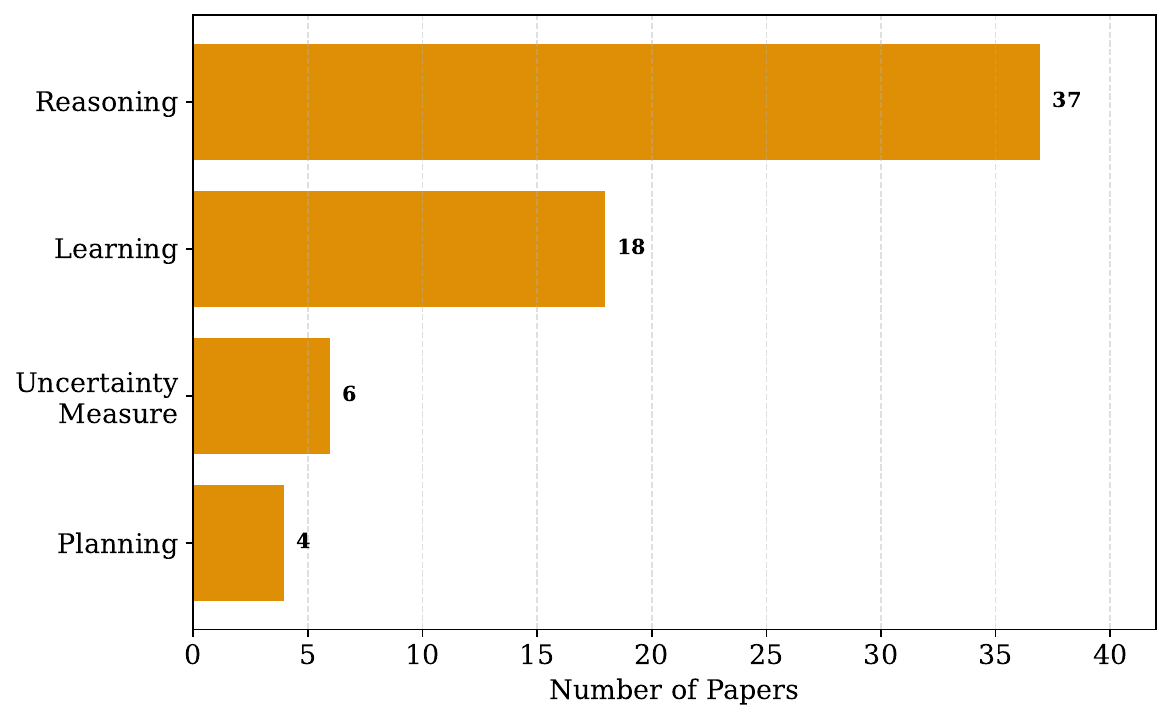}
\caption{Distribution of surveyed papers by primary computational task implemented by hypergame-theoretic models.}
\label{fig:tax_task}
\end{figure}

Finally, we classify hypergame models based on their role in the framework: what task is the approach seeking to solve via hypergame theory? The classification entails planning, uncertainty measure, learning, and reasoning. Models addressing multiple tasks were counted for each class. The overall distribution -- as shown in Figure~\ref{fig:tax_task} -- depicts that the majority of works utilise hypergame-theoretic models for reasoning -- where we include decision-making in general -- or complementing learning algorithms.

\section{Discussion of Survey Results}
\label{sec:results}

\begin{sidewaystable}
\caption{Summary of agent-compatible hypergame paper classification based on model taxonomy and model application details.}\label{summary1}
\begin{tabular*}{\textheight}{@{\extracolsep\fill}lcccccccc}
\toprule%
& \multicolumn{3}{@{}c@{}|}{Hypergame Model Taxonomy}& \multicolumn{5}{@{}c@{}}{Hypergame Application Details} \\\cmidrule{2-4}\cmidrule{5-9}%
Paper & \makecell{Dynamics} & \makecell{Concept} & \multicolumn{1}{c|}{\makecell{Formalization}} & Domain\footnotemark[1] & \makecell{Sub-\\domain\footnotemark[2]} & \makecell{Method} & \makecell{Fidelity} & \makecell{Task}\\
\midrule
\citep{jia2024defense} & Repeated & HNF & HEU-based & CS & DD & Practical & Partial & \makecell{Reasoning,\\Uncertainty\\Measure} \\
\citep{guiterrez2018} & One-shot & MLH & L-th Order & CS & ADM & Theoretical & Complete & Reasoning \\
\citep{bowei2020iobt} & Repeated & MLH & L-th Order & CS & DD & Experimental & Complete & Reasoning \\
\citep{TANG2024103871} & Repeated & MLH & Stackelberg & CS & ADM & Experimental & Complete & Learning \\
\citep{kulkarni2021hypergamegraphs} & Repeated & MLH & Graph-based & CS & ADM & Theoretical & Complete & Reasoning \\
\citep{trencsenyi2025approximatinghumanstrategicreasoning} & One-shot & MLH & \makecell{Flattened\\L-th Order} & SS & TOM & Practical & Complete & Reasoning \\
\citep{kulkarni2021labeling} &  Repeated & MLH & Graph-based & CS & ADM & Theoretical & Complete & Reasoning \\
\citep{kulkarni2020decoyallocation} &  One-shot & MLH & Graph-based & CS & DD & Theoretical & Complete & Reasoning \\
\citep{kulkarni2025coalitiondetection} &  One-shot & MLH & L-th Order & SS & CD & Theoretical & Complete & Learning \\
\citep{lening2023dynamichypergames} &  Repeated & MLH & Graph-based & SS & ADM & Theoretical & Complete & Planning \\
\citep{yuan2024equilibrium} &  Repeated & MLH & HNE-based & GT & DC & Theoretical & Conceptual & Reasoning \\
\citep{gharesifard2012evolution} & Repeated & MLH & Graph-based & GT & EOB & Theoretical & Complete & Learning \\
\citep{gharesifard2010evolution}  & Repeated & MLH & Graph-based & GT & EOB & Theoretical & Complete & Learning \\
\citep{gharesifard2011exploration} & Repeated & MLH & Graph-based & GT & EOB & Theoretical & Complete & \makecell{Learning,\\Uncertainty\\Measure} \\
\citep{wan2021foureyedefensivedeceptionbased} & Repeated & HNF & \makecell{Stackelberg,\\HEU-based} & CS & DD & Practical & Partial & \makecell{Reasoning,\\Uncertainty\\Measure} \\
\citep{anwar2020} & Repeated & HNF & \makecell{HEU-based} & CS & DD & Practical & Partial & \makecell{Reasoning}\\
\citep{li2024hypergameQlearning} & Repeated & MLH & \makecell{Flattened,\\L-th Order} & GT & P-E & Experimental & Complete & \makecell{Reasoning,\\Learning}\\
\citep{house2010hypergame} & Repeated & HNF & HNF & CS & ADM & Experimental & Complete & Learning\\
\citep{thomas2024hypergametheorydecentralizedresource} & Repeated & MLH & Stackelberg & COM & SC & Experimental & Complete & Learning\\
\citep{dharmadhikari2021dronesexploring} & Repeated & MLH & Perceptual & ROB & AV & Practical & Conceptual & Planning\\
\citep{kahn2022traffic} & One-shot & MLH & L-th Order & ROB & CT & Practical & Complete & \makecell{Uncertainty\\Measure}\\
\citep{kulkarni2024integratedresourceallocationstrategy} & Repeated & MLH & Graph-based & CS & DD & Experimental & Complete & Reasoning\\
\citep{shen2025iot} & Repeated & \makecell{MLH,\\HNF} & \makecell{L-th Order,\\HEU-based} & CS & DD & Practical & Complete & Reasoning\\
\citep{thukkaraju2023mia} & Repeated & HNF & HEU-based & CS & ADM & Experimental & Partial & \makecell{Reasoning,\\Learning}\\
\citep{he2022leader} & One-shot & MLH & Stackelberg & ROB & AV & Theoretical & Complete & Reasoning \\
\botrule
\end{tabular*}
\footnotetext[1]{CS: Cybersecurity, SS: Social Simulations, GT: Game Theory, COM: Communications, ROB: Robotics}
\footnotetext[2]{DD: Defensive Deception, ADM: Attack-Defense Modeling, TOM: Thoery of Mind, CD: Coalition Detection, EOB: Evolution of Beliefs, P-E: Pursuit-Evasion, SC: Semantic Communication, AV: Autonomous Vehicles, CT: Control Theory}
\end{sidewaystable}

\begin{sidewaystable}
\caption{(Cont.) Summary of agent-compatible hypergame paper classification based on model taxonomy and model application details.}\label{summary2}
\begin{tabular*}{\textheight}{@{\extracolsep\fill}lcccccccc}
\toprule%
& \multicolumn{3}{@{}c@{}|}{Hypergame Model Taxonomy}& \multicolumn{5}{@{}c@{}}{Hypergame Application Details} \\\cmidrule{2-4}\cmidrule{5-9}%
Paper & \makecell{Dynamics} & \makecell{Concept} & \multicolumn{1}{c|}{\makecell{Formalization}} & Domain\footnotemark[1] & \makecell{Sub-\\domain\footnotemark[2]} & \makecell{Method} & \makecell{Fidelity} & \makecell{Task}\\
\midrule
\citep{bakker2020infomanipulation} & Repeated & MLH & \makecell{L-th Order,\\HNE-based} & CS & ADM & Theoretical & Complete & Reasoning \\
\citep{gharesifard2011learningequilibria} & Repeated & MLH & Graph-based & GT & EOB & Theoretical & Complete & \makecell{Reasoning,\\Learning} \\
\citep{aitchison2021hiddenrolegames} & Repeated & MLH & \makecell{Bayesian\\L-th Order} & MAS & TOM & Practical & Conceptual & \makecell{Reasoning,\\Learning} \\
\citep{bakker2021metagames} & One-shot & MLH & L-th Order & CS & ADM & Theoretical & Complete & Reasoning \\
\citep{Cho2019deceptiongames} & Repeated & HNF & HEU-based & CS & DD & Experimental & Complete & \makecell{Reasoning,\\Uncertainty\\Measure} \\
\citep{kulkarni2019reactivegames} & One-shot & MLH & L-th Order & ROB & CT & Theoretical & Complete & \makecell{Reasoning,\\Planning} \\
\citep{lv2024synthesisforUAV} & Repeated & MLH & Graph-based & ROB & AV & Practical & Complete & Reasoning \\
\citep{chongyang2023stochasticgames} & Repeated & MLH & L-th Order & GT & SG & Experimental & Complete & Planning \\
\citep{udupa2024reactivesynthesissensorrevealing} & Repeated & MLH & Graph-based & CS & DD & Theoretical & Complete & Reasoning \\
\citep{wan2023APT} & Repeated & \makecell{MLH,\\HNF} & \makecell{L-th Order,\\HEU-based} & CS & DD & Practical & Complete & \makecell{Reasoning,\\Learning} \\
\citep{cheng2022stackelberg} & One-shot & MLH & \makecell{Stackelberg,\\HNE-based} & CS & ADM & Theoretical & Complete & Reasoning \\
\citep{gharesifard2014stealthydeception} & Repeated & MLH & Graph-based & CS & ADM & Theoretical & Complete & \makecell{Reasoning,\\Learning} \\
\citep{gharesifard2011stealthy} & Repeated & MLH & Graph-based & GT & EOB & Theoretical & Complete & \makecell{Reasoning,\\Learning}\\
\citep{yoon2025mia} & One-shot & HNF & HEU-based & CS & ADM & Practical & Partial & \makecell{Reasoning,\\Uncertainty\\Measure}\\
\citep{kulkarni2021reachabilitygames} & One-shot & MLH & Graph-based & CS & ADM & Theoretical & Complete & \makecell{Reasoning}\\
\citep{trencsenyi2025influencehumaninspiredagenticsophistication} & One-shot & MLH & \makecell{Flattened\\L-th Order} & SS & TOM & Experimental & Conceptual & \makecell{Reasoning}\\
\citep{hai2023uavcommunication} & One-shot & MLH & Stackelberg & ROB & AV & Theoretical & Complete & \makecell{Reasoning}\\
\citep{tapadhir2020} & Repeated & MLH & L-th Order & CS & TD & Experimental & Complete & \makecell{Reasoning}\\
\citep{song2009fuzzyhypergame} & One-shot & MLH & L-th Order & GT & FUZ & Theoretical & Complete & \makecell{Reasoning}\\

\citep{wan2026cyber} & Repeated & HNF & HEU-based & CS & DD & Practical & Partial & Learning\\
\citep{yoon2026cyber} & Repeated & HNF & HEU-based & CS & ADM & Experimental & Partial & \makecell{Reasoning,\\Learning}\\
\citep{chen2025hypergame} & Repeated & MLH & HNE-based & ROB & AV & Experimental & Partial & \makecell{Reasoning,\\Learning}\\
\citep{ma2025icloud} & One-shot & HNF & HEU-based & CS & DD & Experimental & Partial & \makecell{Reasoning,\\Learning}\\
\citep{yoon2026imia} & One-shot & HNF & HEU-based & CD & ADM & Experimental & Partial & Reasoning\\
\botrule
\end{tabular*}
\footnotetext[1]{CS: Cybersecurity, GT: Game Theory, MAS: Multi-agent Systems, ROB: Robotics, SS: Social Simulations}
\footnotetext[2]{ADM: Attack-Defense Modeling, EOB: Evolution of Beliefs, TOM: Thoery of Mind, DD: Defensive Deception, CT: Control theory, AV: Autonomous Vehicles, TD: Trojan Detection, FUZ: Fuzzy Game Theory}
\end{sidewaystable}

In the preceding sections, we established a comprehensive taxonomy of hypergame models and explored various applications that utilise these models. The full classification of the 44 reviewed papers is documented in Tables~\ref{summary1} and~\ref{summary2}. In this section, we reflect on emerging patterns and structural gaps identified through our systematic review of hypergame-theoretic literature and the classifications disclosed in the previous sections, aiming to contextualise the current state of the field and identify fruitful directions for future research.

\subsection{Hypergame Usage Patterns}

In this section, we analyse emerging patterns in the design and usage of hypergame models across agentic applications. Drawing on the classification dimensions outlined earlier, we identify recurring choices in model structures, domain integration, and computational tasks. These patterns offer practical insights for researchers aiming to deploy hypergame-theoretic mechanisms in dynamic multi-agent systems.

\subsubsection{Overview of Model Usage}
\begin{figure}[htbp]
\centering
\includegraphics[width=\linewidth]{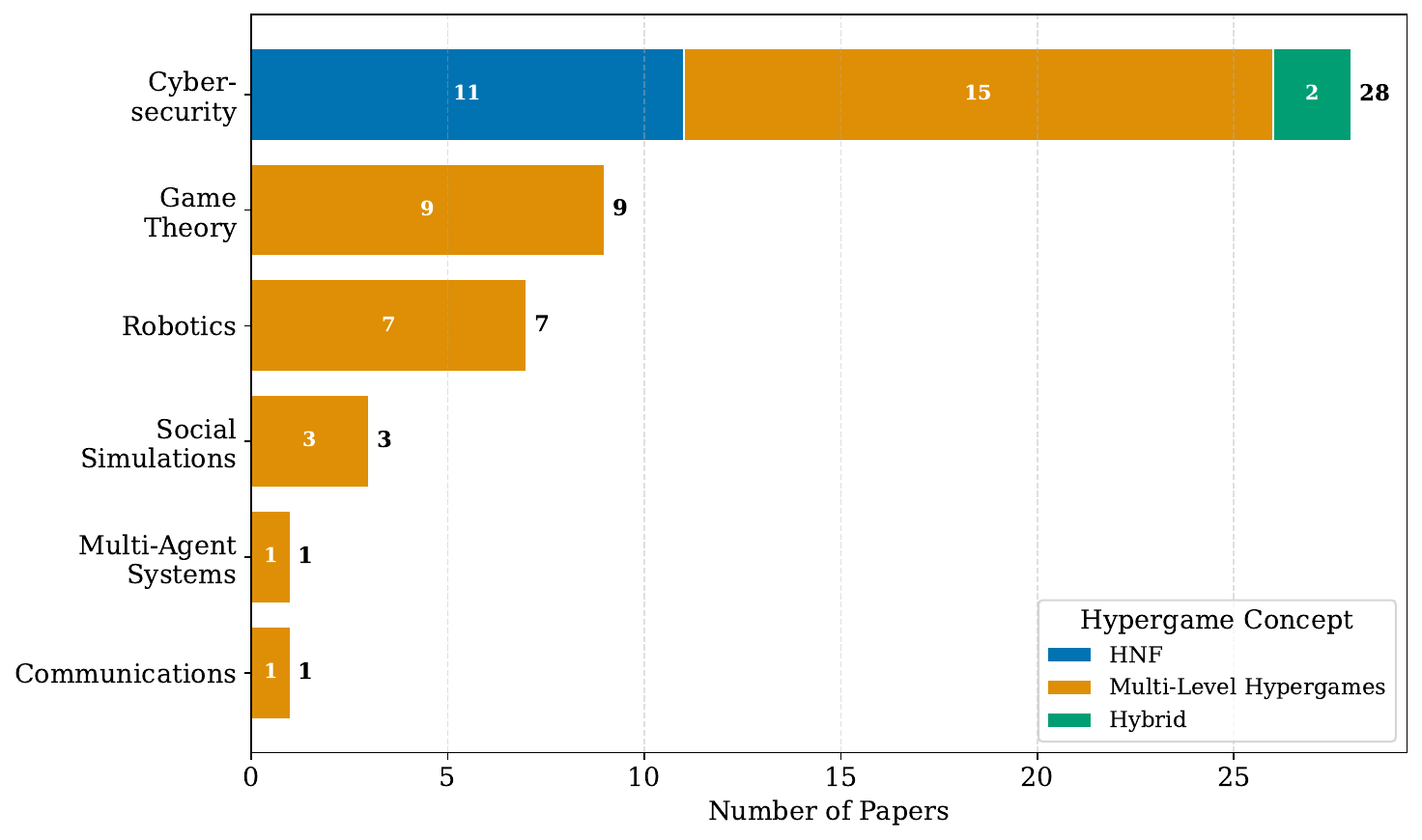}
\caption{Distribution of core hypergame concepts implemented per application domain. \textbf{HNF} entails works inspired by Vane's hypergame normal form~\citep{vane2000thesis}, \textbf{Multi-level Hypergames} are based on Bennet's original concept~\citep{Bennett1980} or Wang et al.'s extended formalism~\citep{Wang1988}. \textbf{Hybrid} approaches use concepts from both main concepts.}
\label{fig:concept-domain}
\end{figure}

An evaluation of the core concept (multi-level hypergame vs HNF) usage across surveyed works at a domain-level decomposition (Figure~\ref{fig:concept-domain}) reveals that multi-level hypergames are the dominant framework across all categories, and only cybersecurity works implement HNF or hybrid models.

\begin{figure}[htbp]
\centering
\includegraphics[width=\linewidth]{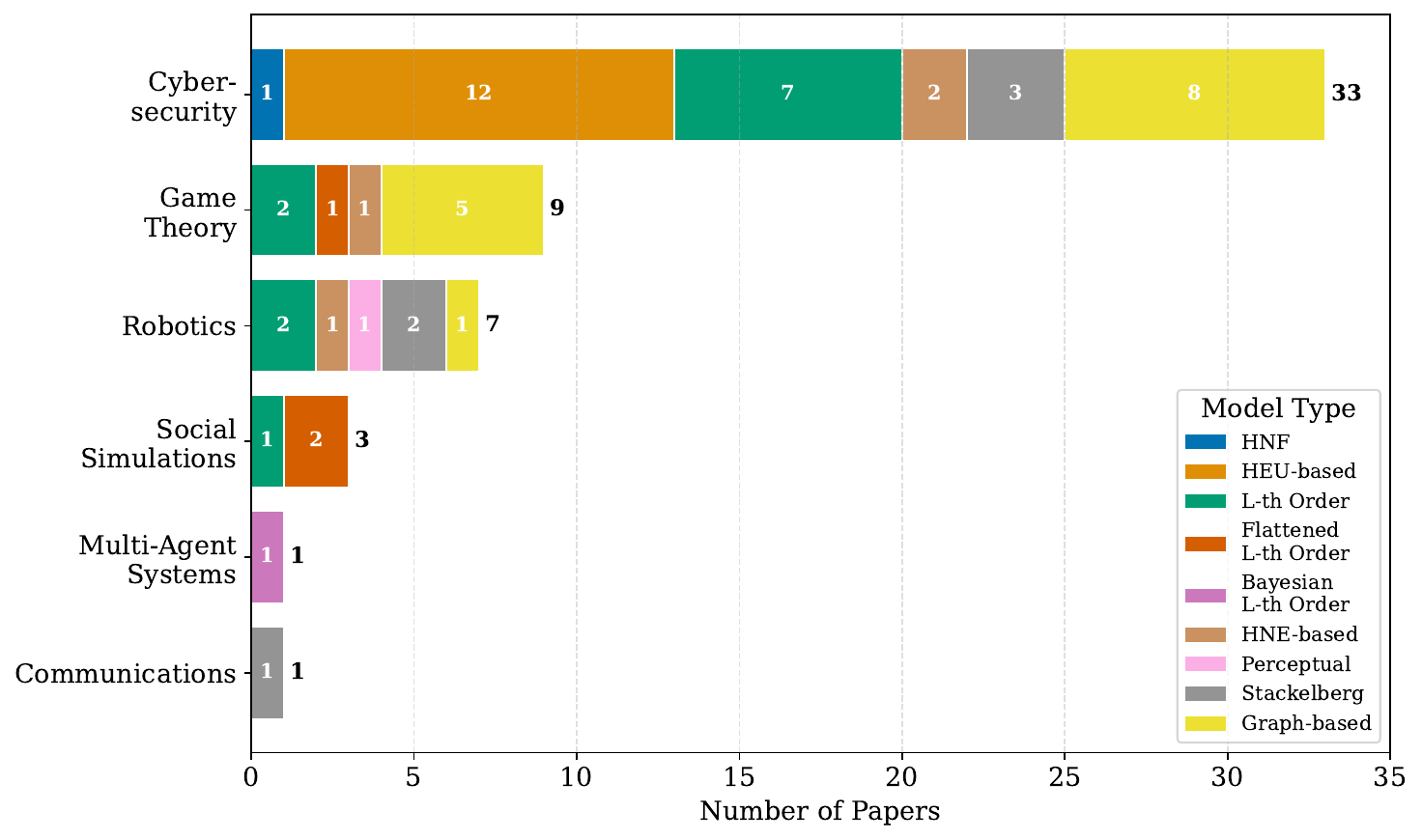}
\caption{Distribution of hypergame model types across application domains. Hybrid models that span multiple categories are counted in each relevant type.}
\label{fig:domain_x_model}
\end{figure}

At a lower-level view, graph-based hypergames and L-th order hierarchical hypergames dominate (Figure~\ref{fig:tax_models}). Graph-based models in particular stand out as the most popular implementation, often supporting spatial planning and adversarial reasoning, especially in cybersecurity and general game-theoretic applications (Figure~\ref{fig:domain_x_model}).

\begin{figure}[htbp]
\centering
\includegraphics[width=\linewidth]{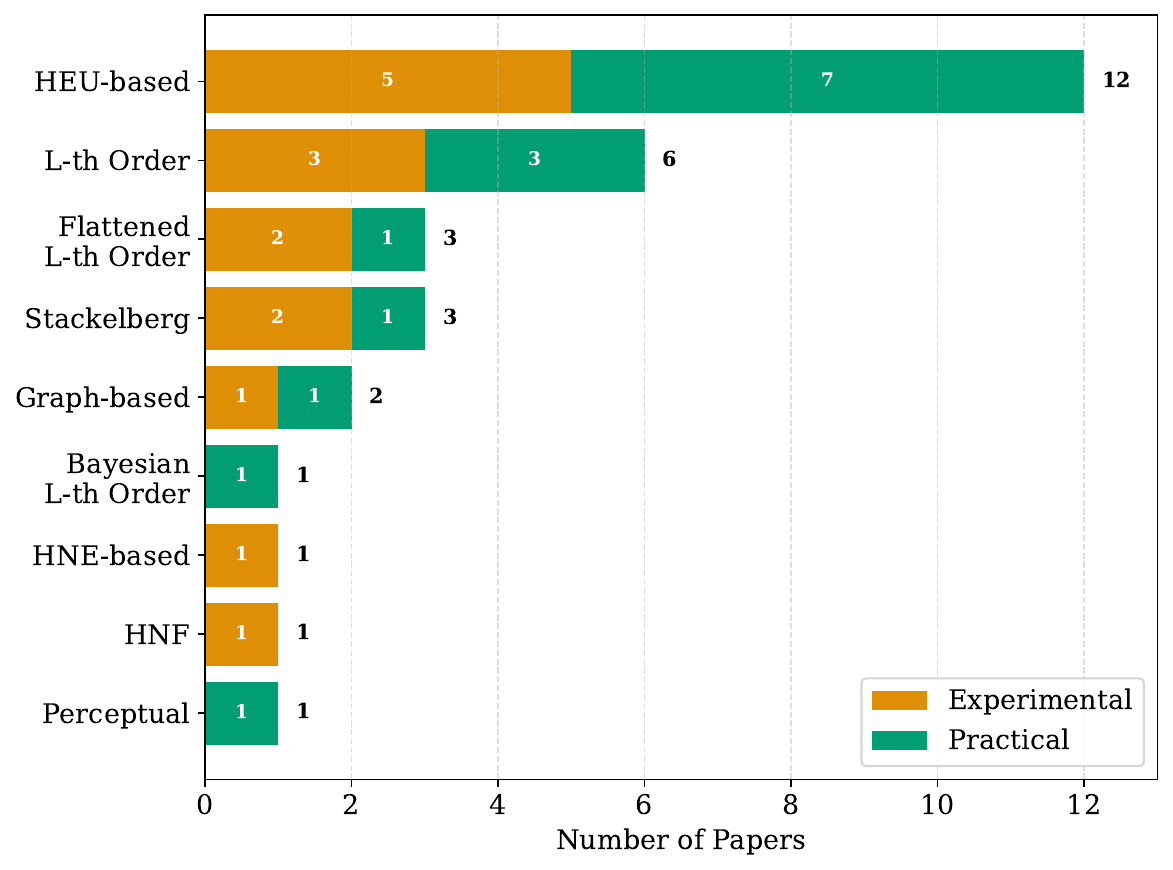}
\caption{Distribution of experimental and practical hypergame applications across model types.}
\label{fig:tax_models_practical}
\end{figure}

\begin{figure}[htbp]
\centering
\includegraphics[width=\linewidth]{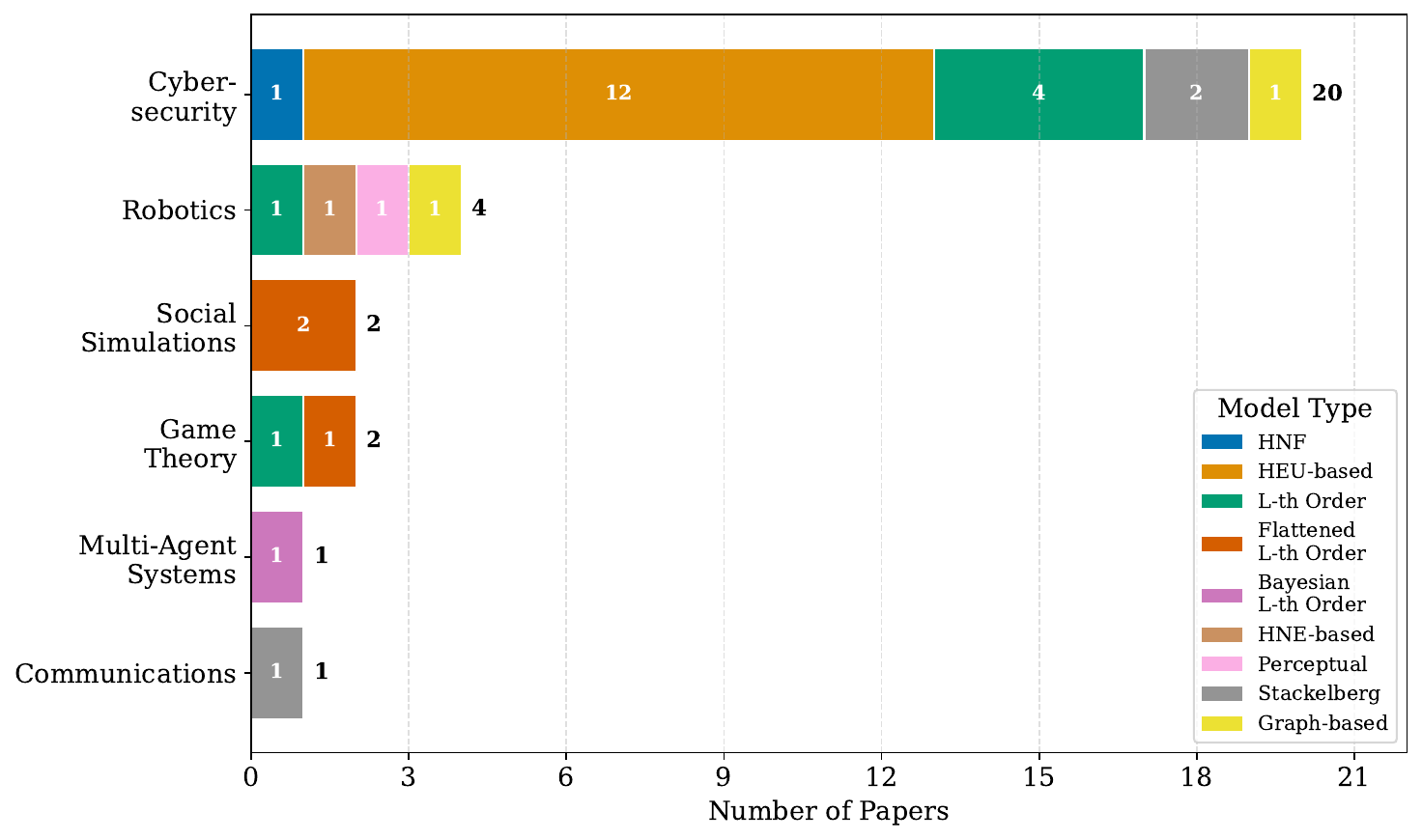}
\caption{Distribution of hypergame model across application domains, omitting theoretical applications.}
\label{fig:domain_x_model_filtered}
\end{figure}

However, when theoretical applications are omitted, and the focus shifts toward experimental and practical implementations closer to deployable multi-agent systems, the distribution changes considerably. As depicted in Figure~\ref{fig:tax_models_practical}, 9 out of the 15 works we classified as practical entail simplified hypergame-theoretic concepts -- perceptual games, flattened L-th order hypergames, and HEU -- rather than integrating and populating models entirely in accordance with complete hypergame analysis frameworks. These findings suggest that while practical, real-world applications can benefit from hypergame-theoretic models, their integration must entail concretisations of the primarily analytical frameworks. Figures~\ref{fig:app_type_main} and~\ref{fig:domain_x_model_filtered} paint a similar picture, where practical applications from domains more aligned with real-world applicability -- cybersecurity, robotics, social simulations, and general MAS -- seem to benefit most from the simplified models.

Meanwhile, HNF has seen limited use in agentic contexts. Only 11 works adopt HNF-based solutions (Figure~\ref{fig:concept-domain}), all of which are situated in the cybersecurity domain. This suggests that HNF’s fixed-matrix structure may be too rigid or low-level for modern agent-based modelling, where higher-order and domain-specific reasoning demands greater representational flexibility.

\subsubsection{Task-Model Alignment}

\begin{figure}[htbp]
\centering
\includegraphics[width=\linewidth]{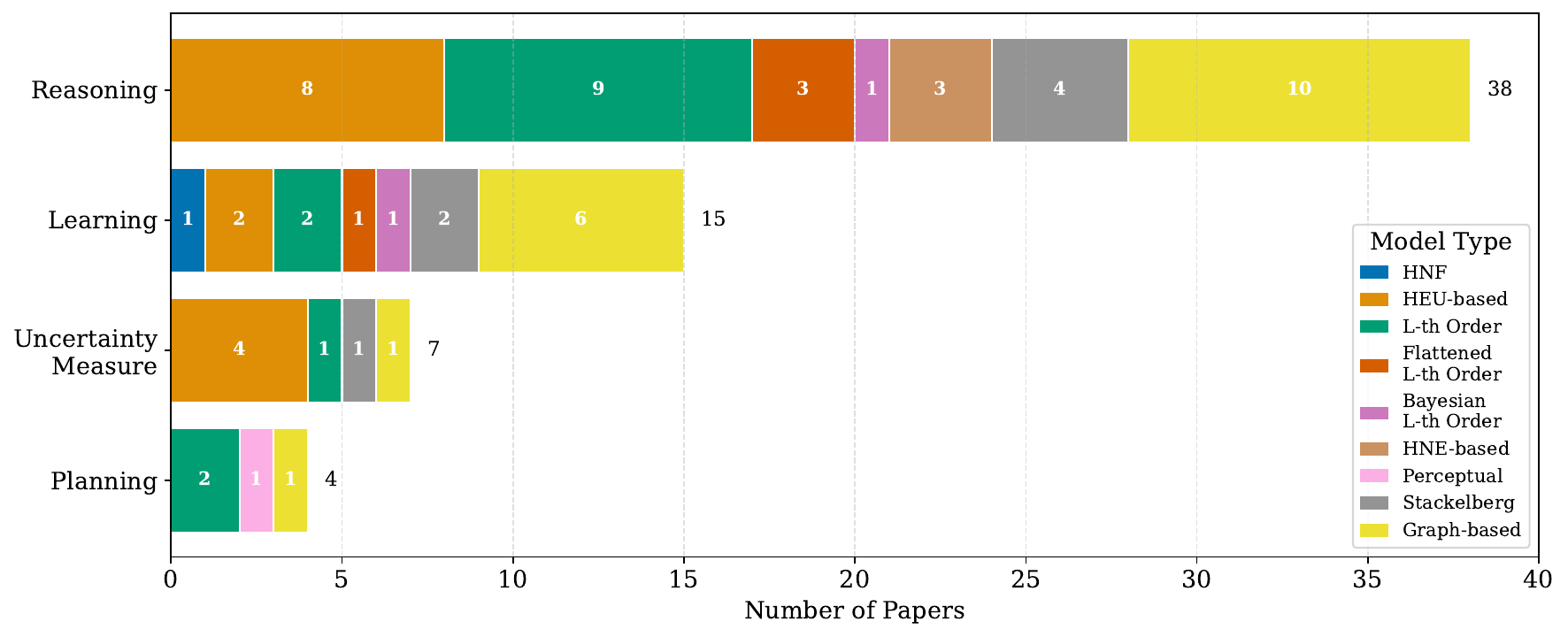}
\caption{Distribution of model types per hypergame task. Note, works can contribute to more than one category.}
\label{fig:tax_model_task}
\end{figure}

As shown in Figure~\ref{fig:tax_model_task}, reasoning is the most common application, with 38 papers using hypergames to support decision-making under misaligned perceptions. Hypergame-based learning tasks appear in about a third of the surveyed papers, most often associated with graph-based and HEU-based models, which offer scalable formulations that integrate well with learning algorithms. Apart from one paper, estimating uncertainty appears as a secondary task, complementing reasoning or learning, with mostly probabilistic, HEU-based approaches. The limited presence of hypergames in high-level planning tasks may reflect the absence of expressive, agent-compatible formalisms akin to epistemic planning languages.

\begin{figure}[htbp]
\centering
\includegraphics[width=\linewidth]{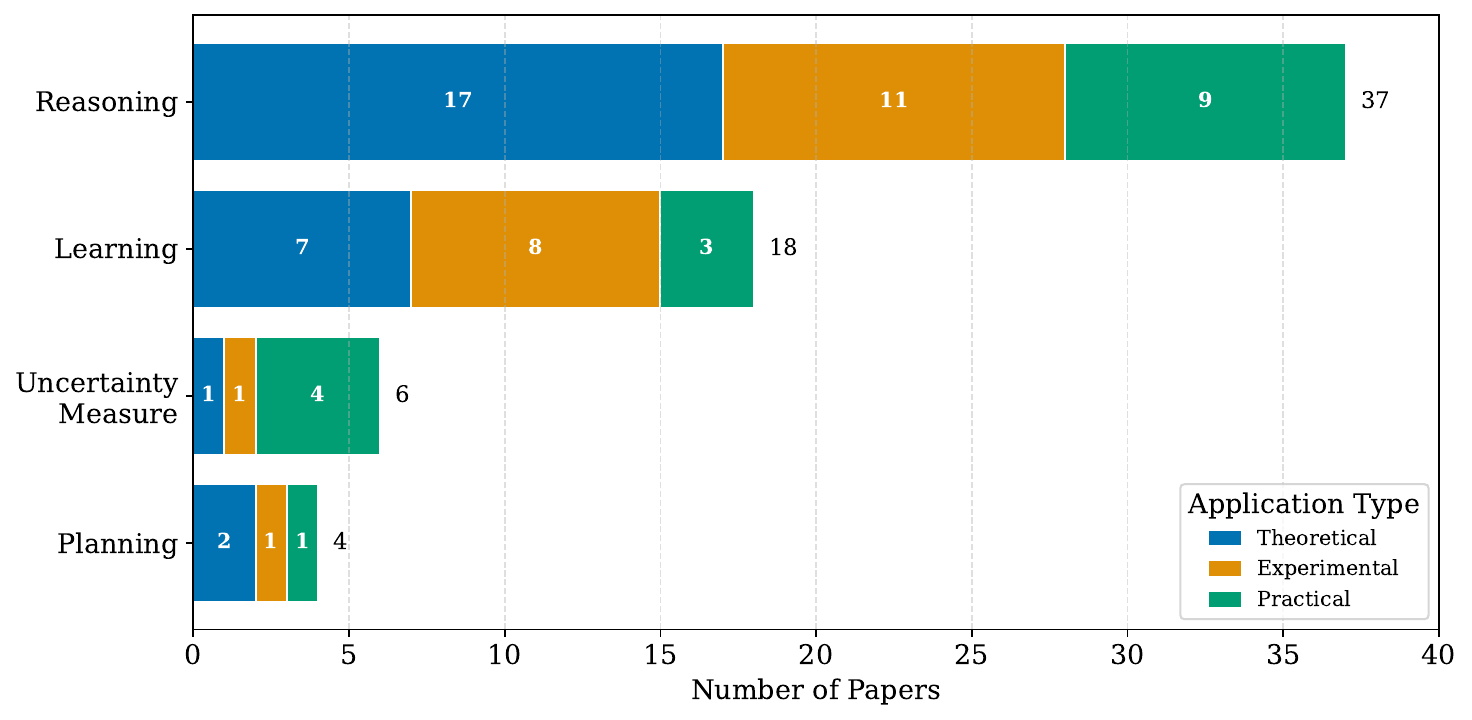}
\caption{Distribution of hypergame task with composition of theoretical, experimental and practical applications.}
\label{fig:task_x_application}
\end{figure}

To complement these findings, we evaluate how different task-based works are composed of theoretical, experimental and practical model applications (Figure~\ref{fig:task_x_application}) and revisit the distribution of hypergame model types per tasks (Figures~\ref{fig:app_type_main},~\ref{fig:domain_x_model_filtered}). While practical applications are the primary focus for deriving hypergame-based uncertainty measures, for planning, learning, and reasoning-related tasks, purely theoretical and more applicable works are split equally.

\subsubsection{Domain-Model Associations}
When decomposed by application domain (Figure~\ref{fig:domain_x_model}), distinct preferences in model types emerge:

\begin{itemize}
    \item Cybersecurity leads in diversity and volume, encompassing both L-th order and HNF-based models, as well as graph-based and Stackelberg hypergames or hypergame equilibrium-based approaches. The prevalence of deception-centric scenarios appears to benefit from nested and multi-layered perception modelling, making hypergames a natural fit.
    \item General game-theoretic works entail mostly theoretical but high-fidelity, complete integrations of hypergame models.
    \item Social simulations and multi-agent system studies predominantly use flattened L-th order and L-th order hypergames, highlighting a focus on recursive reasoning rather than spatial deception. This aligns with their frequent overlap with theory of mind literature and epistemic modelling.
    \item Robotics and communications rely primarily on subjective perceptual representations, which can encode environmental uncertainties and cognitive bias, particularly in navigation and defense tasks.
\end{itemize}

\subsection{Agent-based Hypergame Languages}

While the extended formalisms by~\citep{Wang1988} and~\citep{vane2000thesis} concretised hypergame frameworks and provided guidelines for hypergame analysis, there is no unified formal language for reasoning about hypergames. Motivated by this gap, hypergame markup language (HML) was developed by~\citep{brumley2003hypant} to formally encode hypergame models for his hypergame analysis software: HYPANT. HML supports subjective subgames based on player perception but is limited to static utilities and does not support functional utility definitions or dynamic belief updates. Another hypergame analysis software, HAT~\citep{Gibson2013hypergame}, relies on XML-based structured representation of hypergames. While HAT supports iterative runs and belief updates, it relies on populating Vane's HNF-based structures. The survey by~\citep{kovach2015hypergamereview} also mentions Vane's Security Policy Assistant (SPA), which was developed to assist in sensitive decision-making, such as evaluating foreign disclosure of classified information. SPA is not publicly available, and we could not find sufficient material for a more detailed analysis. Recently,~\citep{trencsenyi2025hypergamerationalisabilitysolvingagent} introduced a declarative, logic-based domain-specific language for encoding hypergame structures and solution concepts. However, its use case remains limited to two-player normal form games and a stateless analytical setting that treats players as static components.

Despite the aforementioned tools and preliminary modelling languages, a general-purpose, agent-based hypergame modelling language or simulation platform has not emerged. While we encountered various works in our survey that integrate hypergame-theoretic models with linear-temporal-logic-based planning, none define a hypergame-specific description language that agents can utilise for deliberation. Building on the logic-grounded foundation offered by~\citep{trencsenyi2025hypergamerationalisabilitysolvingagent}, future hypergame modelling efforts may extend such a formalism beyond two-player, static settings towards dynamic, recursive reasoners. It may take further inspiration from alternative frameworks that address information asymmetry and nested beliefs, such as epistemic logic models that enable dynamic belief updates and planning with iterated knowledge~\citep{baltag2008qualitative,bolander2011epistemic}, GDL-III providing a formal specification language for general game playing with games of incomplete information~\citep{thielscher2017gdl3}, or TOMA extending epistemic logic to formalise the theory of mind~\citep{erdogan2025toma}.

\section{Related Work}
\label{sec:relwork}

While Bennett initially developed hypergame theory as an analytical tool for structuring complex strategic scenarios, subsequent research has explored more dynamic applications. \citep{BENNETT1981shipping} demonstrated the use of hypergame theory in real-time decision-making during a shipping crisis. Later, the development of Vane's HNF provided a more methodological version with potential applications in dynamic, agent-based models. The potential is articulated in his paper addressing decision-theoretic agents~\citep{vane2000dtgt} and his brief, high-level review of hypergames~\citep{vane2006advances}, where the connection to the Observe-Orient-Decide-Act paradigm was also established -- a concept well-aligned with human-cognition-inspired agent theories such as the belief-desire-intention (BDI) architecture~\citep{TWEEDALE2007bdi,Simari2011bdi}. Jointly, including the additional formalism provided by~\citep{Wang1988}, while these advancements of hypergame theory offer a richly formal and generalizable framework, up to this point, this line of research remains unconnected to the agent-based and multi-agent systems literature.

~\citep{kovach2015hypergamereview} presented a broad review of hypergame theory as a framework for modelling conflict, misperception, and deception, particularly in adversarial domains like military planning and cyber defence. Their review emphasises the theoretical basis of hypergames and highlights how layered beliefs and misaligned perceptions create strategic opportunities unavailable in classical game theory or decision theory. However, their review primarily focused on illustrative, mostly analytical examples rather than methodological analyses.

~\citep{zhu2021survey} offered a dual-framework perspective, reviewing defensive deception from both machine learning and game-theoretic viewpoints. Although hypergame theory is featured in their survey through selected analytical and practical examples, the coverage remains limited and application-specific, without substantial linkage to broader multi-agent systems research. Complementing this work,~\citep{khalid2023gtAPT} presented a systematic review of game-theoretic defence models against advanced persistent threats. While not dedicated to hypergames and, as such, lacking formal descriptions of hypergame models, their review acknowledges the value of hypergame theory in modelling adversarial perturbations and covers three additional hypergame-based works on defensive deception. Crook's doctoral dissertation~\citep{crook2024thesis} makes a unique contribution by bridging computability theory and hypergames, exploring how attack graphs in cybersecurity can be recast as hypergames. While preliminary, this work proposes several directions for formalising adaptive, layered-belief hypergames in cybersecurity with links to computable analysis. Finally,~\citep{Fugate_Ferguson-Walter_2019} addressed hypergame theory in their review of game-theoretic models in cyber deception. While their discussion is brief and only on a high level, the paper explicitly recognises that attacker–defender interactions often exhibit asymmetries that traditional games cannot capture. The consideration of hypergames for the adaptability of defender strategies progresses the practical generalizability of hypergame theory.

To our knowledge, no prior work provides a systematic survey of hypergame theory from the perspective of multi-agent systems. Existing reviews typically focus on early decision-theoretic formalisms, defensive deception in cyber-physical systems, or specific ad-hoc and post-hoc case studies. Our survey addresses this gap by:

\begin{itemize}
    \item Providing a rigorous classification framework explicitly tailored to multi-agent systems, emphasising works that implement and apply hypergame theory within agent-compatible contexts;
    \item Presenting a structured taxonomy of hypergame models across multiple dimensions, including formal hypergame structures, agent architectures, domain-specific implementations, and computational tasks;
    \item Identifying key patterns in hypergame integration, such as the prevalence of cybersecurity-related hypergame-theoretic applications, while also highlighting and discussing reasons for the limited use of frameworks like Hypergame Normal Form (HNF) in other domains;
    \item Mapping the explicit relationship between hypergame model types and their deployment in reasoning, planning, learning, and uncertainty management tasks, thereby underscoring the alignment between hypergame formalisms and their intended agentic functionalities;
    \item Demonstrating the evolving landscape of hypergame theory, shifting from historically concentrated theoretical analyses toward more diverse, practical deployments across multiple MAS-relevant domains;
    \item Identifying structural gaps and open research challenges, particularly in hypergame-based epistemic modelling, human–AI interaction, and high-level planning frameworks.
\end{itemize}

By synthesising these insights, our survey not only bridges existing knowledge gaps but also provides a clear roadmap for future hypergame-theoretic research, particularly within dynamic, agent-driven multi-agent systems. The outlined trends and classifications serve as actionable references, encouraging researchers to leverage hypergames in practical MAS applications and inspiring further methodological innovation within the hypergame theory community.

\section{Conclusions}
\label{sec:conc}

In this work, we conducted a systematic review of hypergame theory, focusing on its adaptation and application within multi-agent systems (MAS). We surveyed 49 papers spanning multiple domains and analysed how hypergame models have been integrated into agent-based frameworks, examining trends, formalisations, and application patterns. Our review highlighted both the theoretical strengths of hypergame theory -- particularly its ability to represent nested beliefs, misaligned perceptions, and subjective representations of strategic interactions -- and the practical challenges of deploying such models in dynamic MAS environments.  

Despite early efforts such as HML~\citep{brumley2003hypant} and HAT~\citep{Gibson2013hypergame}, there remains no formal language or simulation framework explicitly designed for modelling agent-based hypergames. Given the identified lack of a unified representational formalism and the need for structured support for recursive belief modelling and dynamic perception updates, future work should prioritise the development of dedicated hypergame description languages tailored to agent-based applications. Inspiration may be drawn from planning frameworks rooted in epistemic logic~\citep{baltag2008qualitative,bolander2011epistemic}, general game-playing languages such as GDL-III~\citep{thielscher2017gdl3}, or formal ToM modelling tools such as TOMA~\citep{erdogan2025toma}. While the minimal hypergame language presented in~\citep{trencsenyi2025hypergamerationalisabilitysolvingagent} can serve as a foundation, its declarative context does not yet support gameplay.

Across the surveyed domains, hypergames have been successfully employed to model nested beliefs, misaligned perceptions, and deception; however, few works explicitly link these models to established cognitive agent architectures. As recursive reasoning becomes increasingly central in agent design, integrating hypergame theory with human-inspired frameworks such as BDI~\citep{TWEEDALE2007bdi,Simari2011bdi} could enrich symbolic reasoning and offer interpretable mechanisms for belief formation and adaptation in dynamic environments. Furthermore, while we encountered only a handful of hypergame-based approaches addressing large language models (LLMs), the growing prevalence of human–agent~\citep{strachan2024testing} and agent–agent misalignment~\citep{Kierans2025misalignment}, along with the application of theory of mind in agentic systems~\citep{bosse2011recursive,rocha2023tomReview,erdogan2025toma}, suggests promising opportunities for hypergame theory to formalise ToM-like models capable of capturing and mitigating agent misalignment.  

Finally, several surveyed works highlighted hybrid reasoning approaches, such as Bayesian $L$-th order hypergames, yet few explored the design space that combines subjective perception modelling with probabilistic inference. Current HNF-based approaches have not yet provided a general solution for dynamically populating the HNF matrix. Future research could explore hybrid agent architectures that integrate hypergame structures for belief representation with probabilistic reasoning modules for decision-making, enabling agents to better navigate uncertain, misperceived, or adversarial contexts.  

We hope that this survey will guide and inspire future researchers seeking to model misperceptions, nested beliefs, and recursive reasoning in multi-agent-based applications. By identifying trends, gaps, and trajectories in the literature, we aim to catalyse the development of hypergame-based AI models that are both theoretically grounded and practically applicable.

\section{Declarations}

\bmhead{Conflict of interest}
The authors declare no conflict of interest.

\bmhead{Funding}
This work was supported by a Leverhulme Trust International Professorship Grant (LIP-2022-001).

\bmhead{Acknowledgement}
The authors would like to express their gratitude to David Levine for his constructive comments. The first author would also like to thank him for his supervision and the fruitful discussion of game-theoretic and hypergame-theoretic materials.

\bmhead{Open Access}
This article is licensed under a Creative Commons Attribution 4.0 International License, which permits use, sharing, adaptation, distribution and reproduction in any medium or format, as long as you give appropriate credit to the original author(s) and the source, provide a link to the Creative Commons licence, and indicate if changes were made. The images or other third party material in this article are included in the article’s Creative Commons licence, unless indicated otherwise in a credit line to the material. If material is not included in the article’s Creative Commons licence and your intended use is not permitted by statutory regulation or exceeds the permitted use, you will need to obtain permission directly from the copyright holder. To view a copy of this licence, visit http://creativecommons.org/licenses/by/4.0/.

\bibliography{references}

\end{document}